\documentclass[conference]{IEEEtran}
\IEEEoverridecommandlockouts
\usepackage{cite}
\usepackage{amsmath,amssymb,amsfonts}
\usepackage{algorithmic}
\usepackage{graphicx}
\usepackage{textcomp}
\usepackage{xcolor}
\def\BibTeX{{\rm B\kern-.05em{\sc i\kern-.025em b}\kern-.08em
    T\kern-.1667em\lower.7ex\hbox{E}\kern-.125emX}}
\usepackage{graphics} 
\usepackage{epsfig} 
\usepackage{mathptmx} 
\usepackage{times} 
\usepackage{amsmath} 
\usepackage{inputenc}
\usepackage{siunitx}
\usepackage{nicefrac}
\usepackage{xcolor}
\usepackage{eqparbox}
\usepackage{tabularx}
\usepackage{todonotes}
\usepackage{csquotes}
\usepackage{url}
\usepackage[caption=false,font=footnotesize]{subfig}
\usepackage{stmaryrd} 
\usepackage{makecell}
\usepackage[acronym, toc, shortcuts, nowarn]{glossaries}
\glsdisablehyper
\setacronymstyle{short-long}
\newacronym{LSTM}{LSTM}{Long Short-Term Memory}
\newacronym{LMU}{LMU}{Legendre Memory Unit}
\newacronym{TCN}{TCN}{Temporal Convolutional Network}
\newacronym{RMSE}{RMSE}{Root-Mean-Square Error}
\newacronym{NGSIM}{NGSIM}{Next Generation Simulation}
\newacronym{LIDAR}{LIDAR}{Light Detection and Ranging}
\newacronym{SPA}{SPA}{Semantic Pointer Architecture}
\newacronym{SNN}{SNN}{Spiking Neural Network}
\newacronym[plural=VSAs, longplural=Vector Symbolic Architectures]{VSA}{VSA}{Vector Symbolic Architecture}
\newacronym{DFT}{DFT}{Discrete Fourier Transform}
\newacronym{IDFT}{IDFT}{Inverse Discrete Fourier Transform}
\newacronym{NEF}{NEF}{Neural Engineering Framework}
\newacronym{AI}{AI}{Artificial Intelligence}
\newacronym{COCO}{MS COCO}{Microsoft Common Objects in COntext}

\newcommand{\norm}[1]{\left\| #1 \right\|}

\begin{document}

\title{The Importance of Balanced Data Sets: Analyzing a Vehicle Trajectory Prediction Model based on Neural Networks and Distributed Representations\\
}

\author{\IEEEauthorblockN{1\textsuperscript{st} Florian Mirus}
\IEEEauthorblockA{\textit{Research, New Technologies, Innovations} \\
\textit{BMW AG}\\
Garching, Germany \\
florian.mirus@bmwgroup.com}
\and
\IEEEauthorblockN{2\textsuperscript{nd} Terrence C. Stewart}
\IEEEauthorblockA{
\textit{Applied Brain Research Inc.}\\
Waterloo, Ontario, Canada \\
terry.stewart@appliedbrainresearch.com}
\and
\IEEEauthorblockN{3\textsuperscript{rd} J\"org Conradt}
\IEEEauthorblockA{\textit{Dep. of Comp. Science and Technology} \\
\textit{KTH Royal Institute of Technology}\\
Stockholm, Sweden \\
conr@kth.se}
}

\maketitle

\begin{abstract}
    Predicting future behavior of other traffic participants is an essential task that needs to be solved by automated vehicles and human drivers alike to achieve safe and situation-aware driving.
    Modern approaches to vehicles trajectory prediction typically rely on data-driven models like neural networks, in particular \acp{LSTM}, achieving promising results.
    However, the question of optimal composition of the underlying training data has received less attention.
    In this paper, we expand on previous work on vehicle trajectory prediction based on neural network models employing distributed representations to encode automotive scenes in a semantic vector substrate.
    We analyze the influence of variations in the training data on the performance of our prediction models.
    Thereby, we show that the models employing our semantic vector representation outperform the numerical model when trained on an adequate data set and thereby, that the composition of training data in vehicle trajectory prediction is crucial for successful training.
    We conduct our analysis on challenging real-world driving data.
\end{abstract}


\section{Introduction}%
\label{sec:introduction}

The race to autonomous driving is currently one of the main forces for pushing research forward in the automotive domain.
With highly automated vehicle prototypes gradually making their way to our public roads and fully-automated driving on the horizon, it seems to be a matter of time until we see robot taxis or cars navigating us through urban traffic or heavy stop-and-go on highways.
One major reason for this development in recent years is the rapid progress of \ac{AI}, especially the success of deep learning, which has shown remarkable results in tasks essential for automated driving like object detection, classification \cite{Ciresan2012} and control \cite{Bojarski2016}.

\begin{figure}[t]
    \centering
    \includegraphics[width=1.\linewidth]{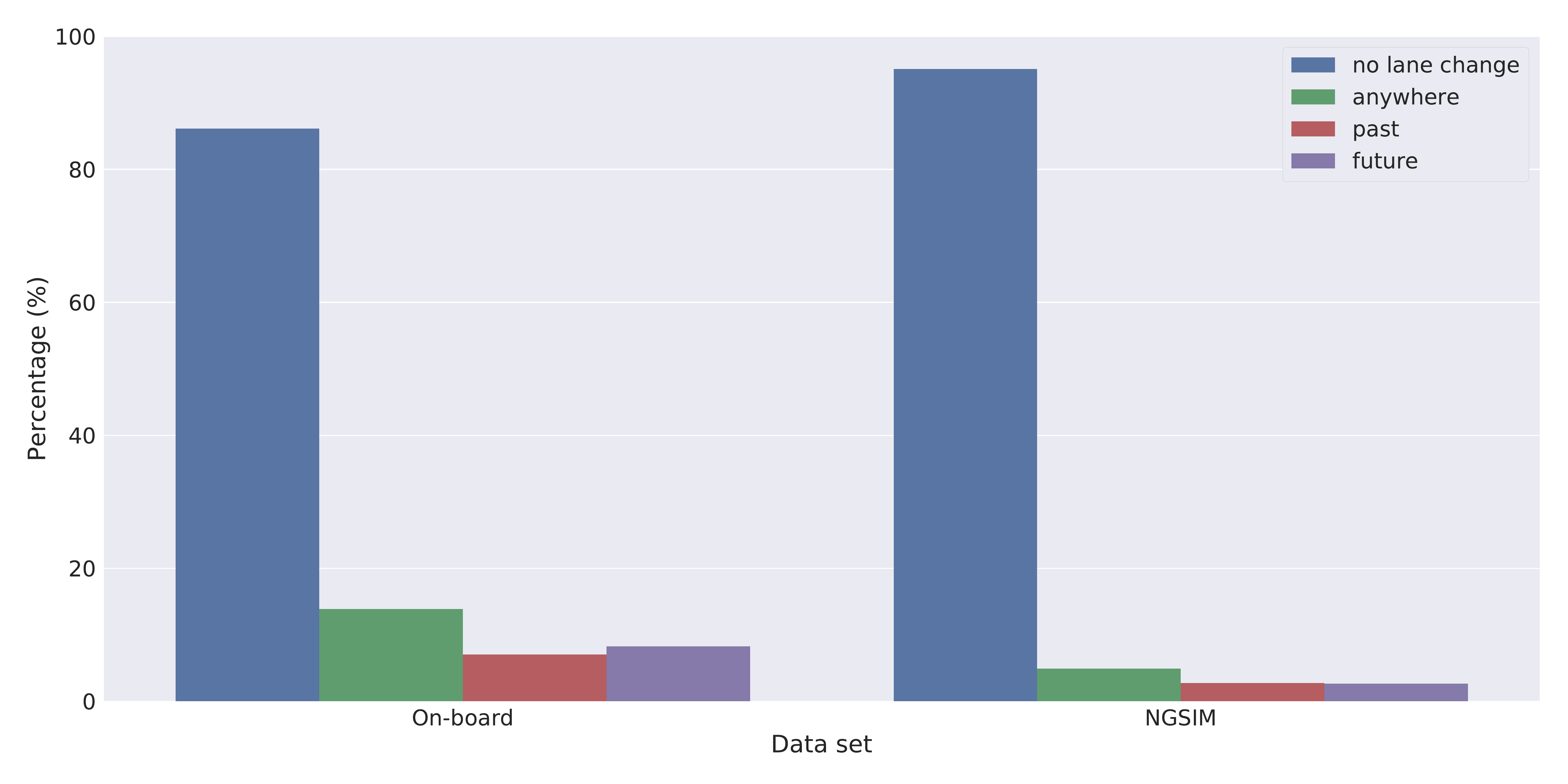}
    \caption{Visualization of the composition of both our data sets regarding lane changes of the target vehicle.}
    \label{fig:data_set_lane_change_distribution}
\end{figure}

Predicting future behavior and positions of other traffic participants from observations is a cornerstone for successful collision avoidance and thus safe motion planning, and needs to be solved by human drivers and automated vehicles alike to reach their desired goal.
Motion prediction for intelligent vehicles in general has seen extensive research in recent years \cite{Lawitzky2013, Lefevre2014, Polychronopoulos2007, Schmuedderich2015}.
Lef{\`e}vre et al. \cite{Lefevre2014} classify such prediction approaches into three categories, namely \emph{physics-based}, \emph{maneuver-based} and \emph{interaction-aware}, depending on their level of abstraction.
\emph{Physics-based} and \emph{maneuver-based} motion models consider the law of physics and the intended driving maneuver respectively as the only influencing factors for future vehicle motion and ignore inter-dependencies between the motion of different vehicles.
There exists a growing number of different \emph{interaction-aware} approaches to account for those dependencies and mutual influences between traffic participants or, more generally, agents in the scene.
Probabilistic models like costmaps \cite{Bahram2016} account for physical constraints on the movements of the other vehicles.
Classification approaches categorize and represent scenes in a hierarchy \cite{Bonnin2012} based on the most generic ones to predict behavior for a variety of different situations.

\begin{figure*}[t]
    \centering
    \includegraphics[width=0.95\linewidth]{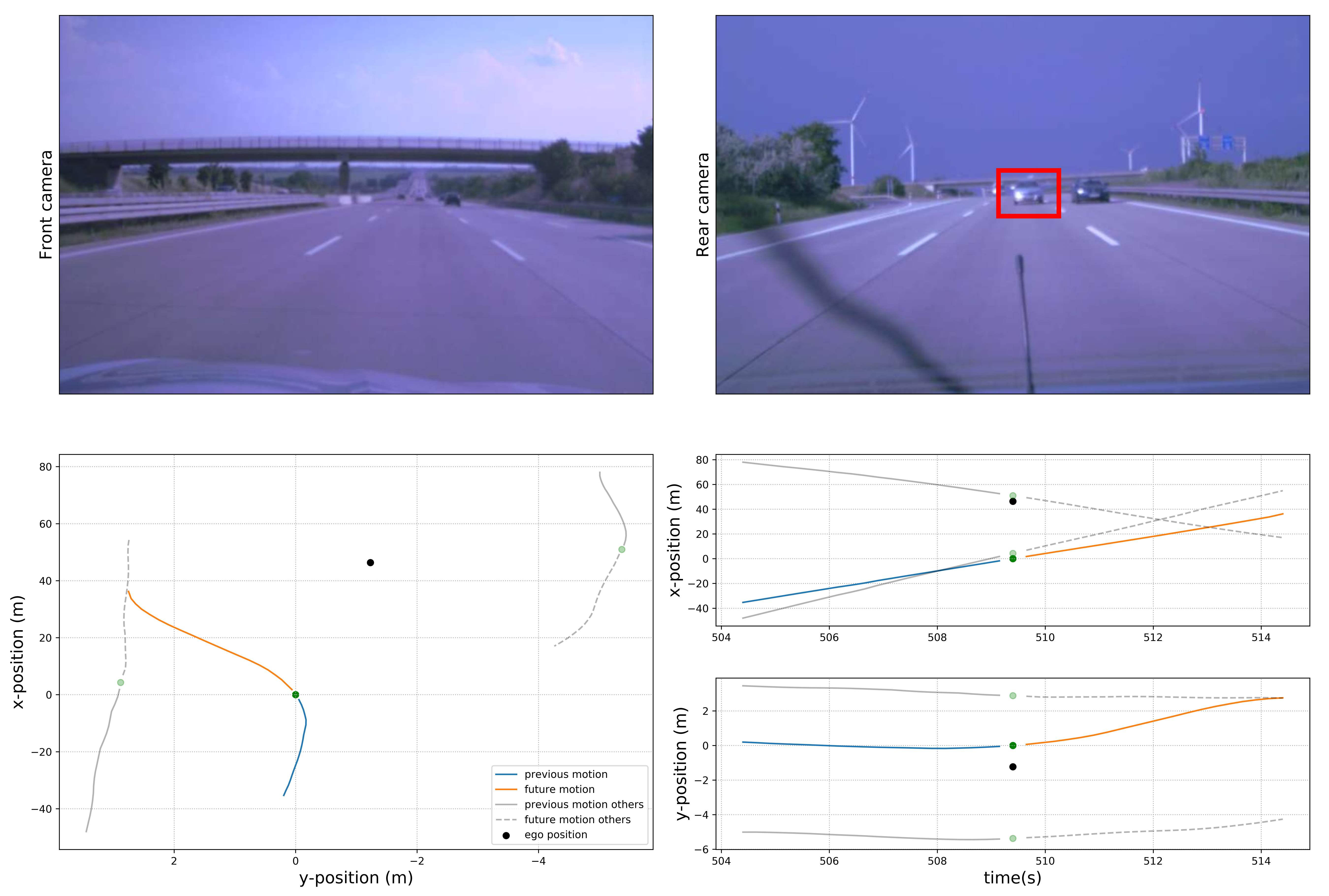}
    \caption{Data visualization of one data sample from the \emph{On-board} data set $D_1$ containing a future lane change of the target vehicle.
        The dots in the left plot indicate the position of the vehicles and color-code the vehicle type (red=motorcycle, green=car, blue=truck, black=ego-vehicle), blue and orange lines show past and future motion of the target vehicle whereas gray lines depict the other vehicles' motion.
        The images in the top row show raw images recorded using the ego-vehicle's front and rear camera with the target vehicle highlighted by a red bounding box.}
    \label{fig:on_board_lane_change_example_with_imgs}
    \vspace{-0.3cm}
\end{figure*}
Data-driven approaches to behavior prediction mainly rely on \acf{LSTM} neural network architectures \cite{Hochreiter1997}, which have proven to be a powerful tool for sequential data analysis.
Alahi et al. \cite{Alahi2016} model interactions in the learning network architecture by introducing so-called social-pooling layers to connect several \acp{LSTM} each predicting the distribution of the trajectory position of one agent at a time.
Deo and Trivedi \cite{Deo2018a} adapted the combination of \ac{LSTM} networks for encoding vehicle trajectories and (convolutional) social-pooling layers to account for interactions to vehicle prediction in highway situations.
Altche and de La Fortelle \cite{Altche2018} use a \ac{LSTM} network as well, but they account for interactions by including distances between the target vehicle and other agents directly in the training data rather than adapting the network architecture.
A similar approach is proposed in \cite{Deo2018}, but it combines \ac{LSTM} networks with an additional maneuver classification network to predict future vehicle motion.

However, for any model employing a data-driven learning approach, the underlying data set used for training is of crucial importance for the model to successfully learn and, more importantly, generalize beyond unseen examples. 
Compared to the field of computer vision with many publicly available data sets such as Imagenet \cite{Deng2009}, \ac{COCO} \cite{Lin2014} or, specifically in automotive context, the KITTI \cite{Geiger2013a}, Cityscapes \cite{Cordts2016} or ApolloScapes \cite{Huang2018} data sets for training and evaluating data-driven models, there is a lack of publicly available data sets for vehicle trajectory prediction.
The currently most widely used data set for trajectory prediction \cite{Altche2018, Deo2018, Deo2018a, Mirus2019b} is the \emph{\ac{NGSIM}} data set \cite{NGSIM-US101}, which was originally created for traffic flow analysis and contains only examples from highway driving.
Furthermore, this data set suffers from a strong imbalance towards straight driving compared to more \enquote{interesting} driving maneuvers such as lane changes (cf. Fig.~\ref{fig:data_set_lane_change_distribution}).
The recently proposed INTERACTION data set \cite{Zhan2019} as well as the trajectory data set from ApolloScapes \cite{Huang2018} are a first step to tackle this issue.

\subsection{Contribution}%
\label{subsec:contribution}

\begin{figure*}[t]
    \centering
    \includegraphics[width=0.99\linewidth]{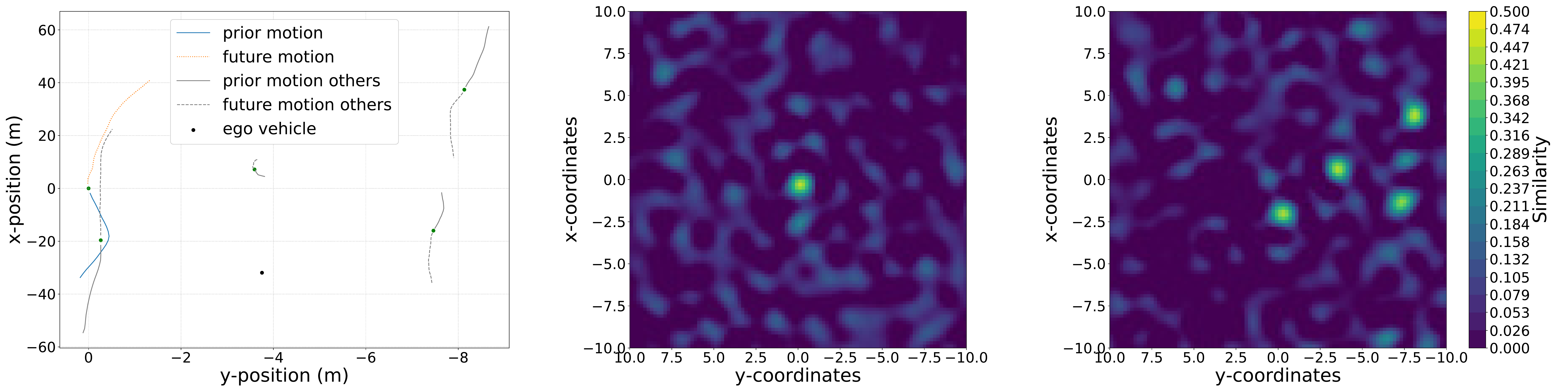}
    \caption{Visualization of the convolutive power representation for \num{512}-dimensional vectors.
        The left plot depicts a scene from the \emph{On-board} data set, while the middle and right plots visualize the similarity between the representation vector of that scene and auxiliary comparison vectors created from a sequence of discrete values as heat map for the target vehicle (middle) and other cars (right).
    }
    \label{fig:spa_power_representation}
    \vspace{-0.3cm}
\end{figure*}

In this work, we adopt \ac{LSTM}-based vehicle prediction models as well as an environment representation based on distributed representations from previous work \cite{Mirus2019b}.
We found that vehicle prediction models could benefit from our vector representation in certain situations such as, for instance, crowded driving situations with multiple vehicles in close surrounding of the target- and/or ego-vehicle.
In this paper, we extend that work by analyzing the influence of the data set used for training on the models' performance.
Here, we focus on comparing models trained on the complete set of samples or solely on training samples containing a lane change performed by the target vehicle (cf. Fig.~\ref{fig:on_board_lane_change_example_with_imgs}). 
Following this approach, our contribution is two-fold: firstly, we are able to show that models employing our semantic vector representation proposed in \cite{Mirus2019b}, which encodes information about vehicle interactions directly within the semantics of the vectors themselves, perform better than models using a simple numerical encoding of the input data without access to mutual interactions between agents when trained on lane change samples.
Secondly, we find that the composition of the data used for training models for vehicle trajectory prediction is crucial for successful learning and should be considered in more detail in future research.

\section{Materials and Methods}%
\label{sec:materials_and_methods}

\subsection{Data and Preprocessing}%
\label{subsec:data_and_preprocessing}

In this work, we use two different data sets for training and evaluation of our system, which we refer to as \emph{On-board} or $D_1$, which is a proprietary data set containing real-world data gathered during test drives mainly on highways in southern Germany, and \emph{\ac{NGSIM}} or $D_2$, which is a publicly available data set recorded using external cameras observing a segment of approximately \SI{640}{\meter} length with \num{6} lanes on the US-101 freeway in Los Angeles, California.
Both data sets contain object-lists with a variety of features such as position, velocity and acceleration but also object type probabilities and lane information.
The \emph{On-board} data set contains \num{3891} vehicles, which yield a total length of roughly \SI{28.3}{\hour} when adding up the time each individual vehicle is visible, whereas the \emph{\ac{NGSIM}} data set contains \num{5930} vehicles and therefore a total time of roughly \SI{91.3}{\hour}.
For training and evaluating our model, we split both data sets into training $T_i \subset D_i$ and validation data $V_i \subset D_i$ containing \SI{90}{\percent} and \SI{10}{\percent} of the objects respectively to avoid testing our models on vehicles they have been trained with.
Although we used both data sets in previous work \cite{Mirus2019b}, we conduct our analysis regarding the composition of the training data in this paper, however, solely on the \emph{On-board} data set and only state a similar or even more imbalanced composition of the data set with respect to lane change maneuvers for the \emph{\ac{NGSIM}} data set (cf. Fig.~\ref{fig:data_set_lane_change_distribution})

\subsection{Data set composition}%
\label{subsec:data_set_composition}

In this section, we analyze the composition of our data sets regarding the amount of \enquote{interesting} behavior of the target vehicle.
Both, the \emph{On-board} and \emph{\ac{NGSIM}} data set consist of mainly highway driving, where we would expect mainly straight driving with the most interesting situations being the target vehicle, i.e., the vehicle whose motion we aim to predict, performing a lane change.
Hence, we are interested in the amount of situations where the target vehicle actually performs a lane change and how much more prominent normal straight driving is in both our data sets.
For the \emph{On-board} data set, we have information about the current lane as well as the distance to the lane borders estimated from the ego-vehicle's cameras available for all vehicles.
The \emph{\ac{NGSIM}} data set contains information about the current lane for each vehicle extracted from the external camera's video footage.
Thus, the selection process for the examples containing a lane change is straightforward for both data sets.
Figure~\ref{fig:on_board_lane_change_example_with_imgs} shows one data sample from the \emph{On-board} data set containing a lane change performed by the target vehicle in its future motion to be predicted.
Comparing this example to the one shown in Fig.~\ref{fig:spa_power_representation}, which shows mainly straight driving for all vehicles present in the scene, we observe that a lane change mainly influences the vehicle's motion in lateral ($y$) direction.

Figure~\ref{fig:data_set_lane_change_distribution} shows the amount of situations where the target vehicle performs a lane change in comparison to the amount of situations where no such behavior occurs for both, the \emph{On-board} and \emph{\ac{NGSIM}} data set.
For the \emph{On-board} data set, in \SI{86.1}{\percent} of all data samples the target vehicle does not perform a lane change, i.e., only \SI{13.8}{\percent} of all data samples contain a lane change performed by the target vehicle.
We further distinguish between lane changes performed during the trajectory history, i.e., the past \SI{5}{\second} before the current time step (labeled as \emph{past} in Fig.~\ref{fig:data_set_lane_change_distribution}) and lane changes that are performed in the future, i.e., during the future \SI{5}{\second} from the current time step (labeled as \emph{future} in Fig.~\ref{fig:data_set_lane_change_distribution}).
We consider the lane changes in the future part of data samples to be the most interesting and challenging ones, since any model making predictions about the future trajectory needs to be able to anticipate these lane changes.
For the \emph{On-board} data set, \SI{7}{\percent} of all data samples contain a lane change in the trajectory history, while \SI{8.2}{\percent} of the samples contain a future lane change performed by the target vehicle.
In comparison to the \SI{86.1}{\percent} of data samples not containing a lane change, the amount of samples with interesting behavior other than straight driving within the data set is significantly less present.
For the \emph{\ac{NGSIM}} data set, the discrepancy between the amount of samples without the target vehicle performing a lane change and the number of samples containing a lane change is even more significant.
The percentage of samples without a target vehicle lane change is \SI{95.1}{\percent} while only \SI{4.9}{\percent} of the samples contain a lane change performed by the target vehicle at all.
The amount of samples containing a future lane change performed by the target vehicle is only \SI{2.6}{\percent} of all samples in the \emph{\ac{NGSIM}} data set.
Hence, there is a significant imbalance in both data sets between examples containing mainly straight driving by the target vehicle, namely \SI{86.1}{\percent} and \SI{95.1}{\percent} of all samples in the \emph{On-board} and \emph{\ac{NGSIM}} data set respectively, where most likely already simple prediction approaches are able to achieve reasonable results.

\subsection{Convolutive vector-power}%
\label{subsec:convolutive_vector_power}

The \ac{SPA} \cite{Eliasmith2013} is one special case of \aclp{VSA} \cite{Gayler2003}, a family of modeling approaches based on high-dimensional vector representations.
Such vectors are one example of distributed representations in the sense that information is captured over all dimensions of the vector instead of one single number, which allows to encode both, symbol-like and numerical structures in a similar and unified way.
Additionally, the architectures' algebraic operations allow manipulation and combination of represented entities into structured representations.
In our work, atomic vectors are picked from the real-valued unit sphere, the dot product serves as a measure of similarity and the algebraic operations are component-wise vector addition $\oplus$ and circular convolution $\otimes$.
In this work, we make use of the fact that for any two vectors $\mathbf{v},\mathbf{w}$, we can write
\begin{equation}
  \mathbf{v} \otimes \mathbf{w} = \acs{IDFT} \left(\acs{DFT}(\mathbf{v}) \odot \acs{DFT}(\mathbf{w})\right),
  \label{eq:conv_dft}
\end{equation}
where $\odot$ denotes element-wise multiplication, \acs{DFT} and \acs{IDFT} denote the \acl{DFT} and \acl{IDFT} respectively.

Using Eq.~\eqref{eq:conv_dft}, we define the \emph{convolutive power} of a vector $\mathbf{v}$ by an exponent $p \in \mathbb{R}$ as
\begin{equation}
  \mathbf{v}^{p} := \Re\left(\acs{IDFT} \left(\left(\acs{DFT}_{j}\left(\mathbf{v}\right)^{p}\right)_{j=0}^{D-1}\right)\right),
  \label{eq:conv_power}
\end{equation}
where $\Re$ denotes the real part of a complex number.
Furthermore, we call a vector $\mathbf{u}$ \emph{unitary}, if $\norm{\mathbf{v}} = \norm{\mathbf{v} \otimes \mathbf{u}}$ for any other $\mathbf{v}$ (see \cite[Sec. 3.6.3 and 3.6.5]{Plate1994} for more details on the convolutive power and unitary vectors) .

\subsection{Vector representation}%
\label{subsec:vector_representation}

In this paper, we adopt the vector representation of automotive scenes for trajectory prediction \cite{Mirus2019b} with the goal to evaluate the performance of the models proposed in \cite{Mirus2019b} when trained and evaluated on different data set variations.
We assign a random ID-vector to each category of dynamic objects (e.g., car, motorcycle, truck) as well as random unitary vectors $\mathbf{X}$ and $\mathbf{Y}$ to encode spatial positions.
Let $(x, y)$ denote the position of the target vehicle and $(x_{obj}, y_{obj})$ the positions of all other visible objects closer than \SI{40}{\meter} to the target (to avoid accumulation of noise), we encapsulate this information in a scene vector
\begin{align}
\label{eq:scene_vector}
\mathbf{S} &= \mathbf{TARGET}\otimes \mathbf{TYPE}_{target} \otimes \mathbf{X}^{x} \otimes \mathbf{Y}^{y} \nonumber \\ 
           &\oplus \sum_{obj} \mathbf{TYPE}_{obj} \otimes \mathbf{X}^{x_{obj}} \otimes \mathbf{Y}^{y_{obj}},
\end{align}
where $\mathbf{TARGET}$ denotes an additional ID-vector chosen at random to indicate the target object (relevant for trajectory prediction).
Figure~\ref{fig:spa_power_representation} visualizes one example scene from the \emph{On-board} data set and its representation vector queried for the target and other cars.

\subsection{Prediction models}%
\label{subsec:prediction_models}

\begin{figure*}[th!]
    \centering
    \subfloat[\label{subfig:rmse_large_all}]{%
        \includegraphics[width=0.25\linewidth]{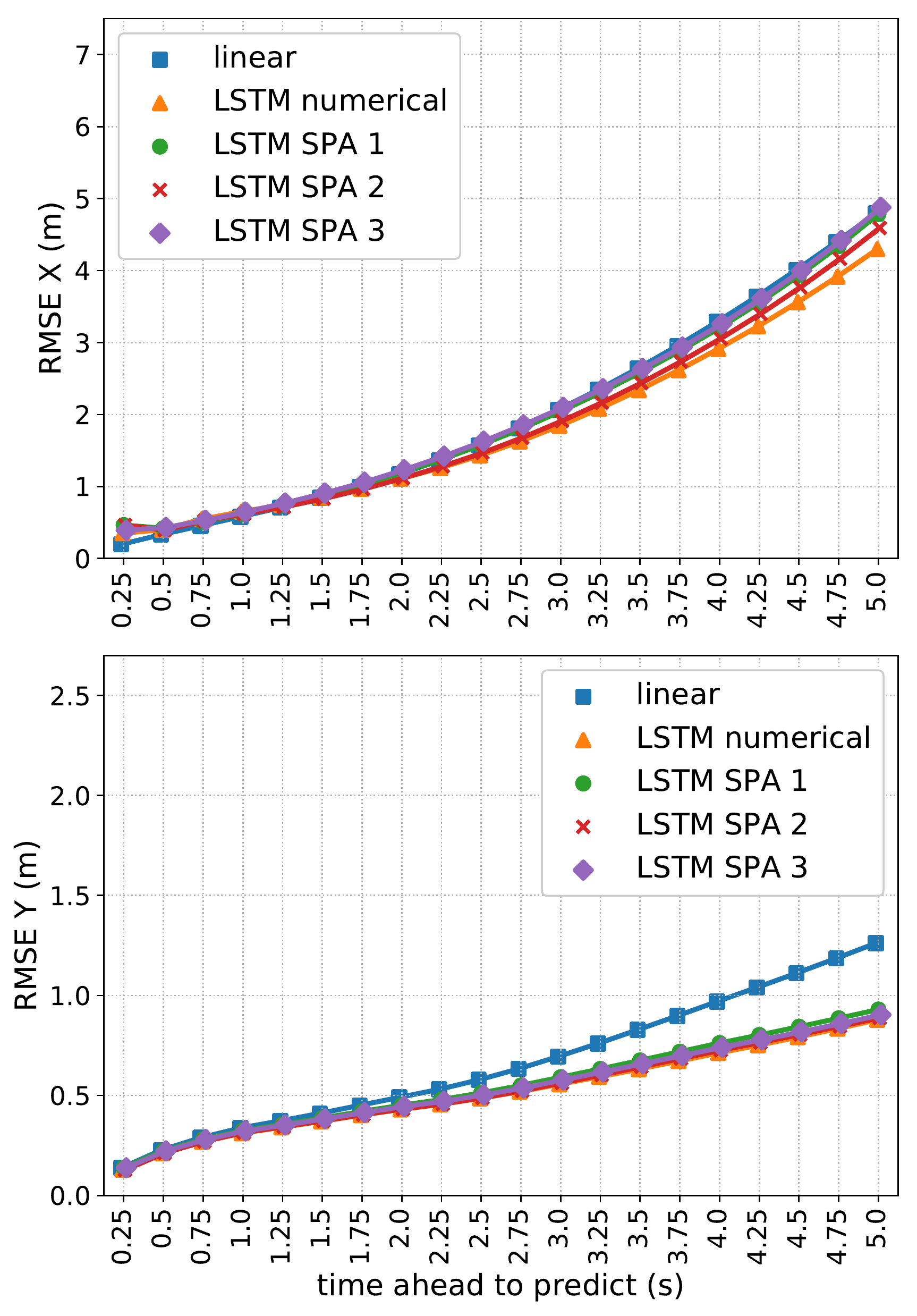}
    }
    \subfloat[\label{subfig:rmse_large_all_trained_on_lc}]{%
        \includegraphics[width=0.25\linewidth]{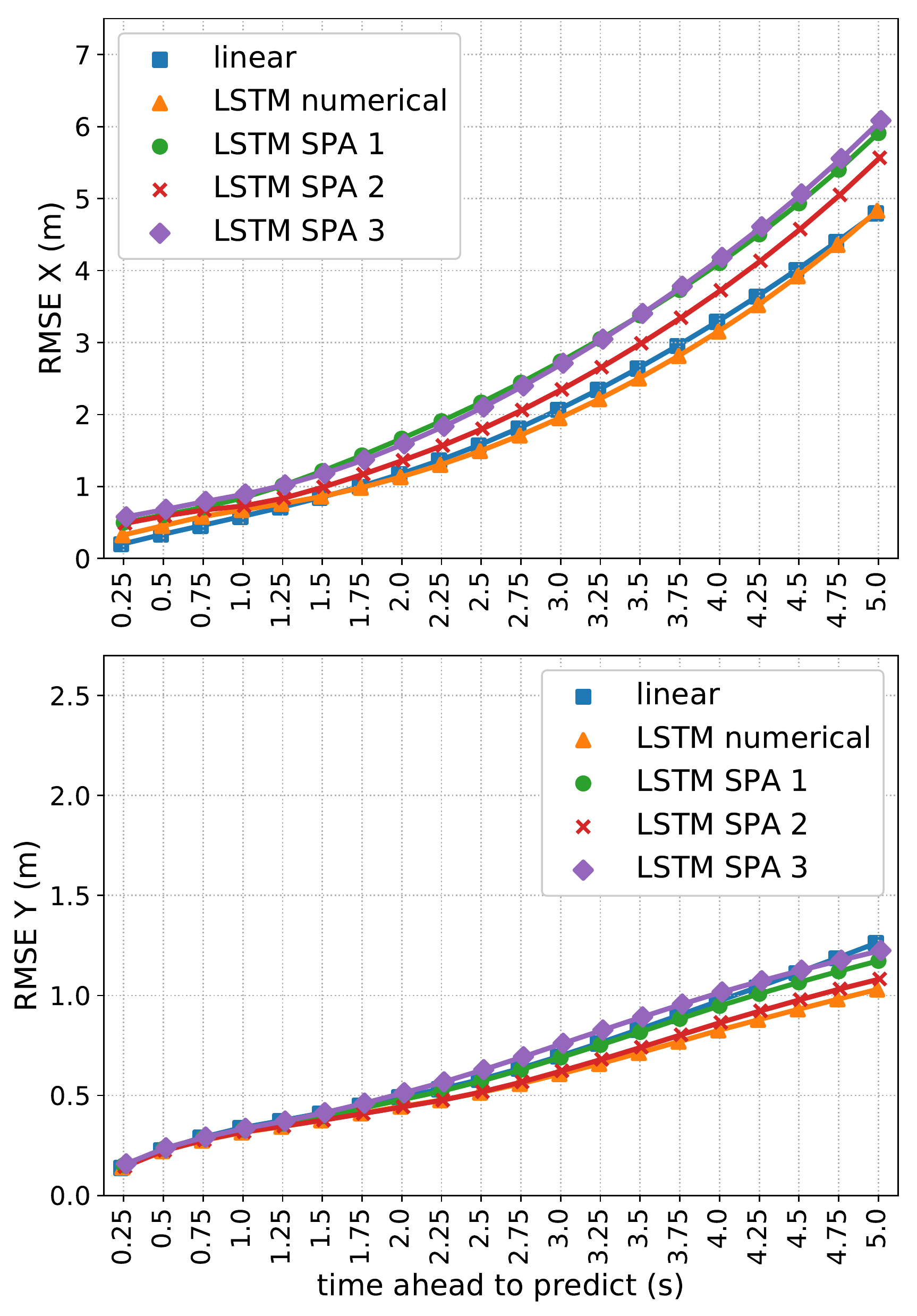}
    }
    \subfloat[\label{subfig:rmse_large_all_lc_only}]{%
        \includegraphics[width=0.25\linewidth]{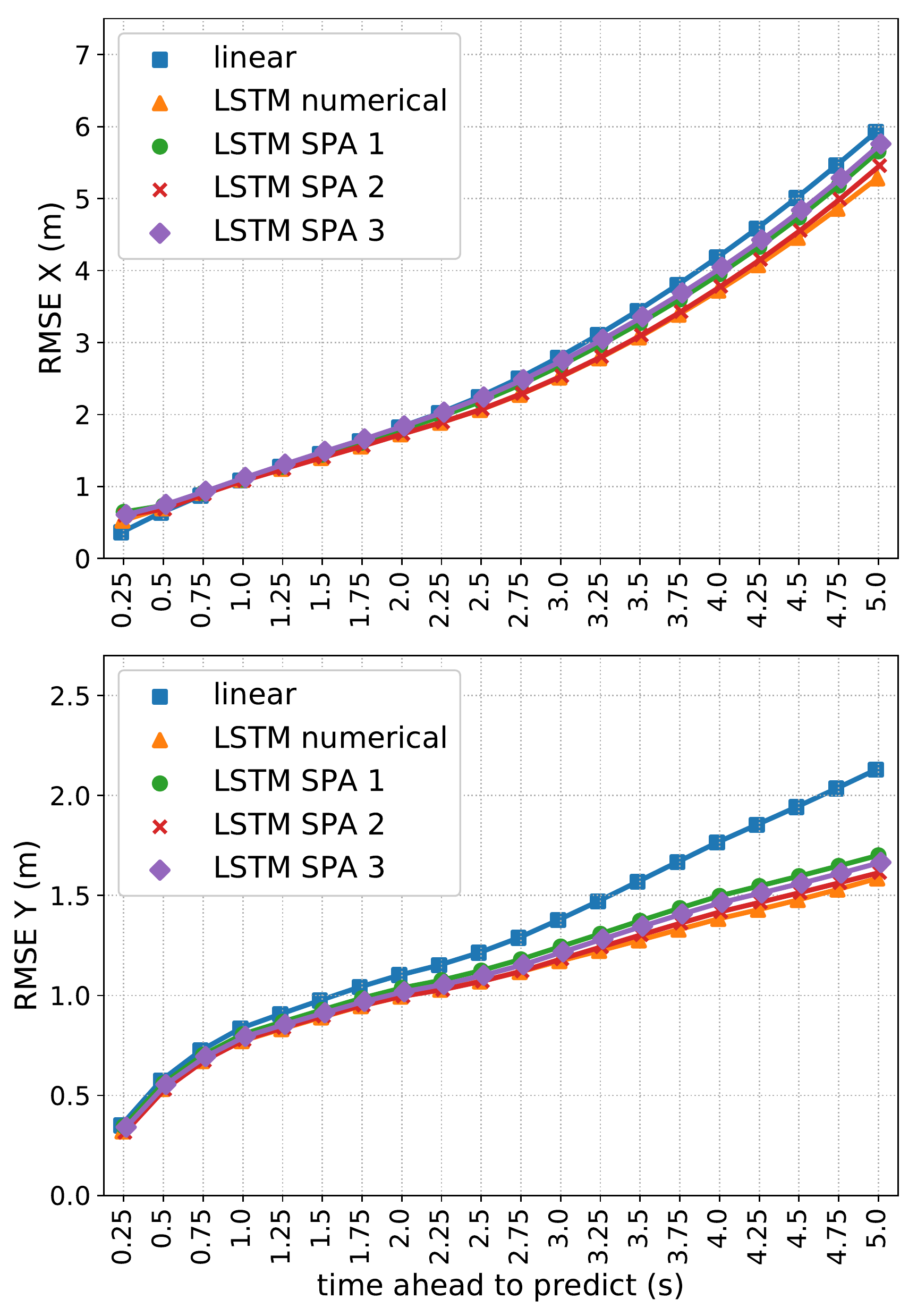}
    }
    \subfloat[\label{subfig:rmse_large_all_lc_only_trained_on_lc}]{%
        \includegraphics[width=0.25\linewidth]{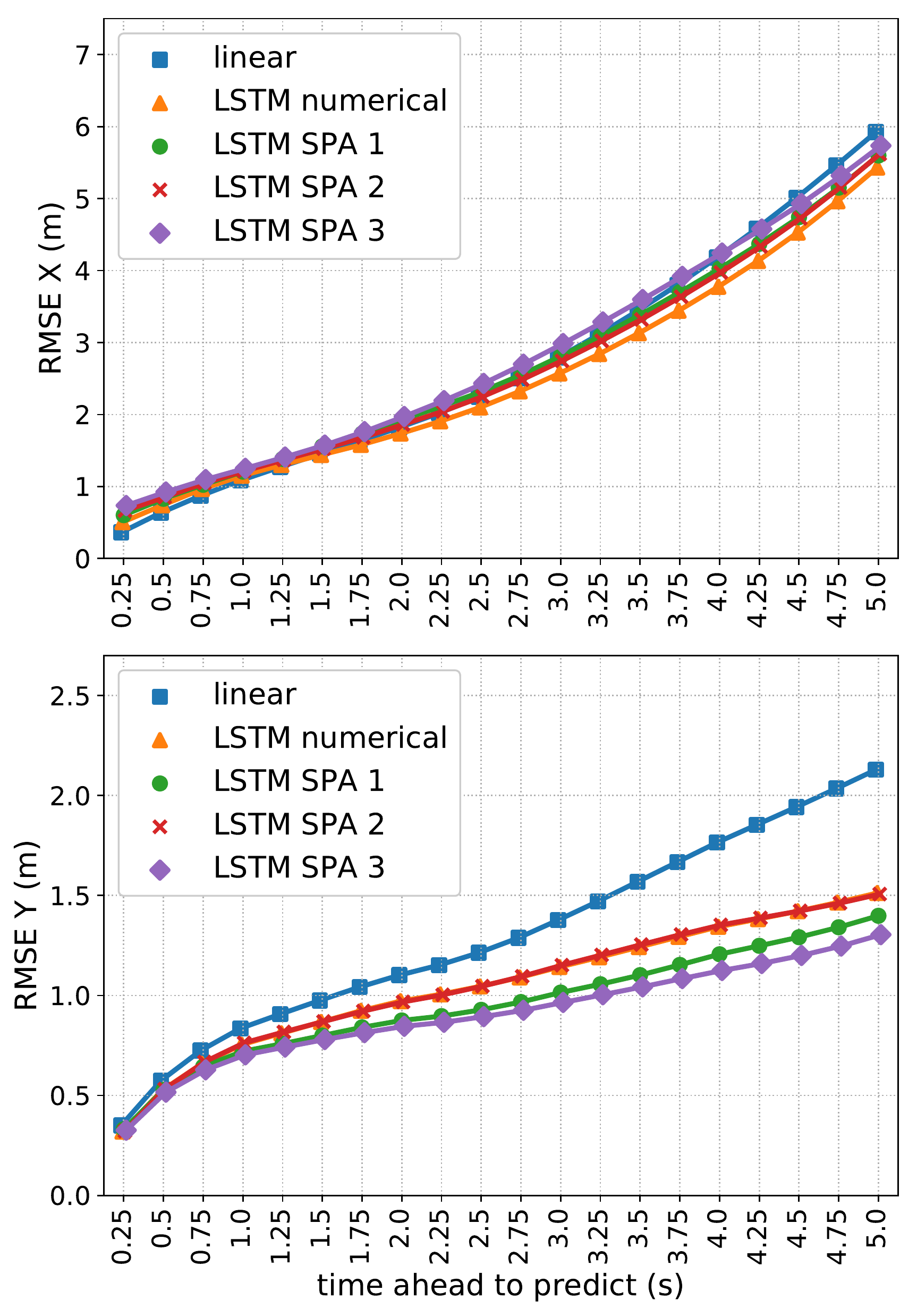}
    }\\
    \subfloat[\label{subfig:rmse_large_subset}]{%
        \includegraphics[width=0.25\linewidth]{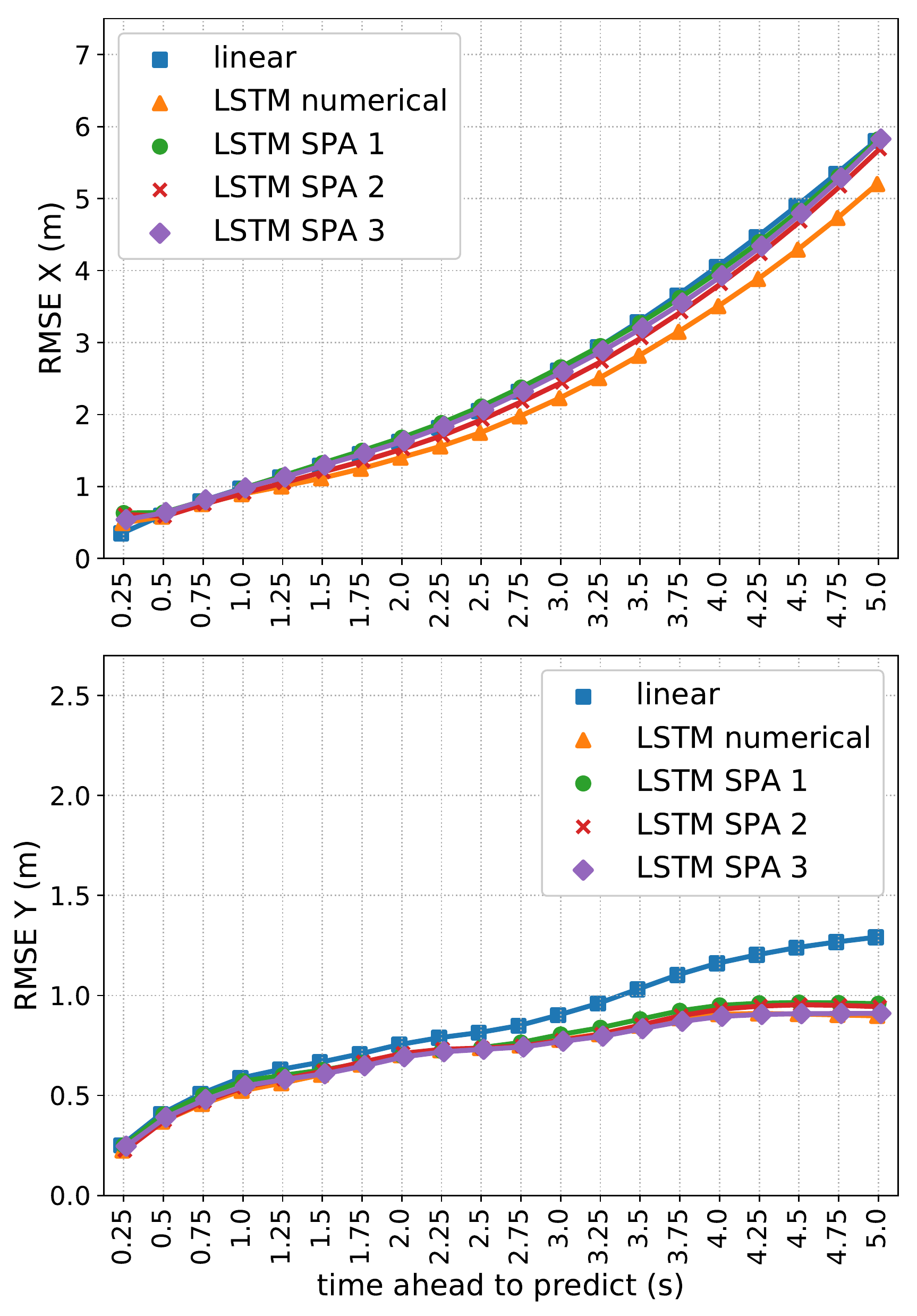}
    }
    \subfloat[\label{subfig:rmse_large_subset_trained_on_lc}]{%
        \includegraphics[width=0.25\linewidth]{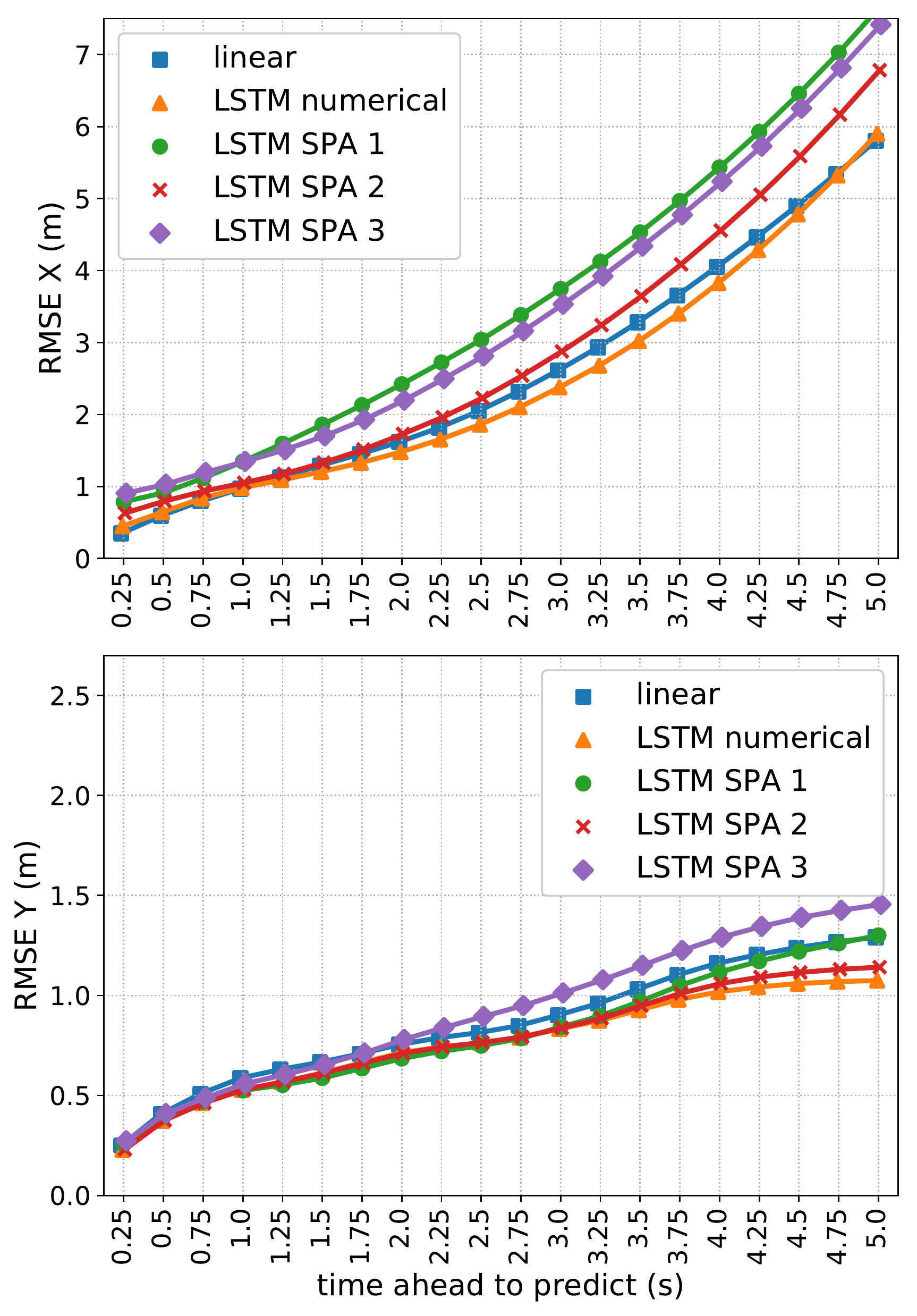}
    }
    \subfloat[\label{subfig:rmse_large_subset_lc_only}]{%
        \includegraphics[width=0.25\linewidth]{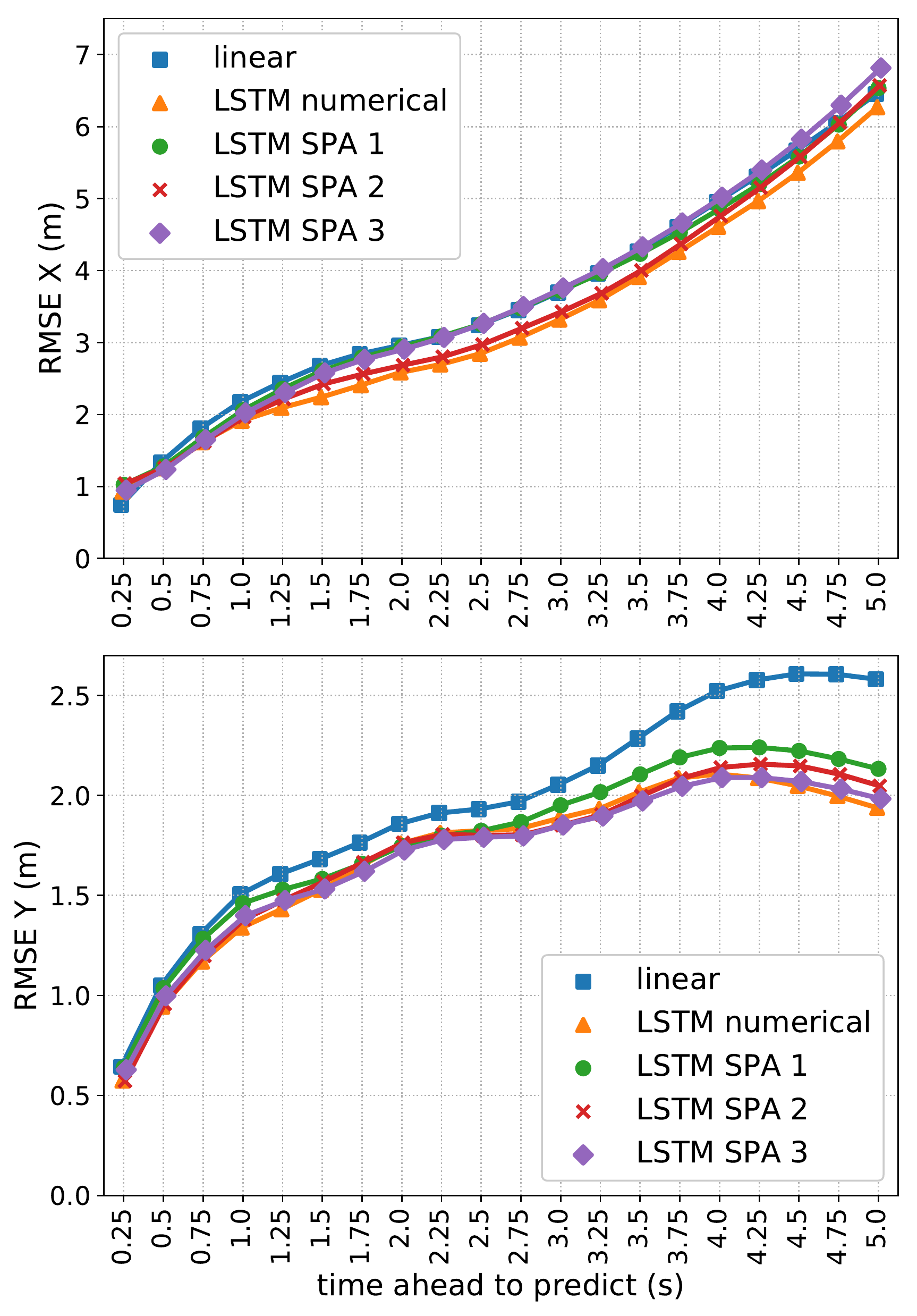}
    }
    \subfloat[\label{subfig:rmse_large_subset_lc_only_trained_on_lc}]{%
        \includegraphics[width=0.25\linewidth]{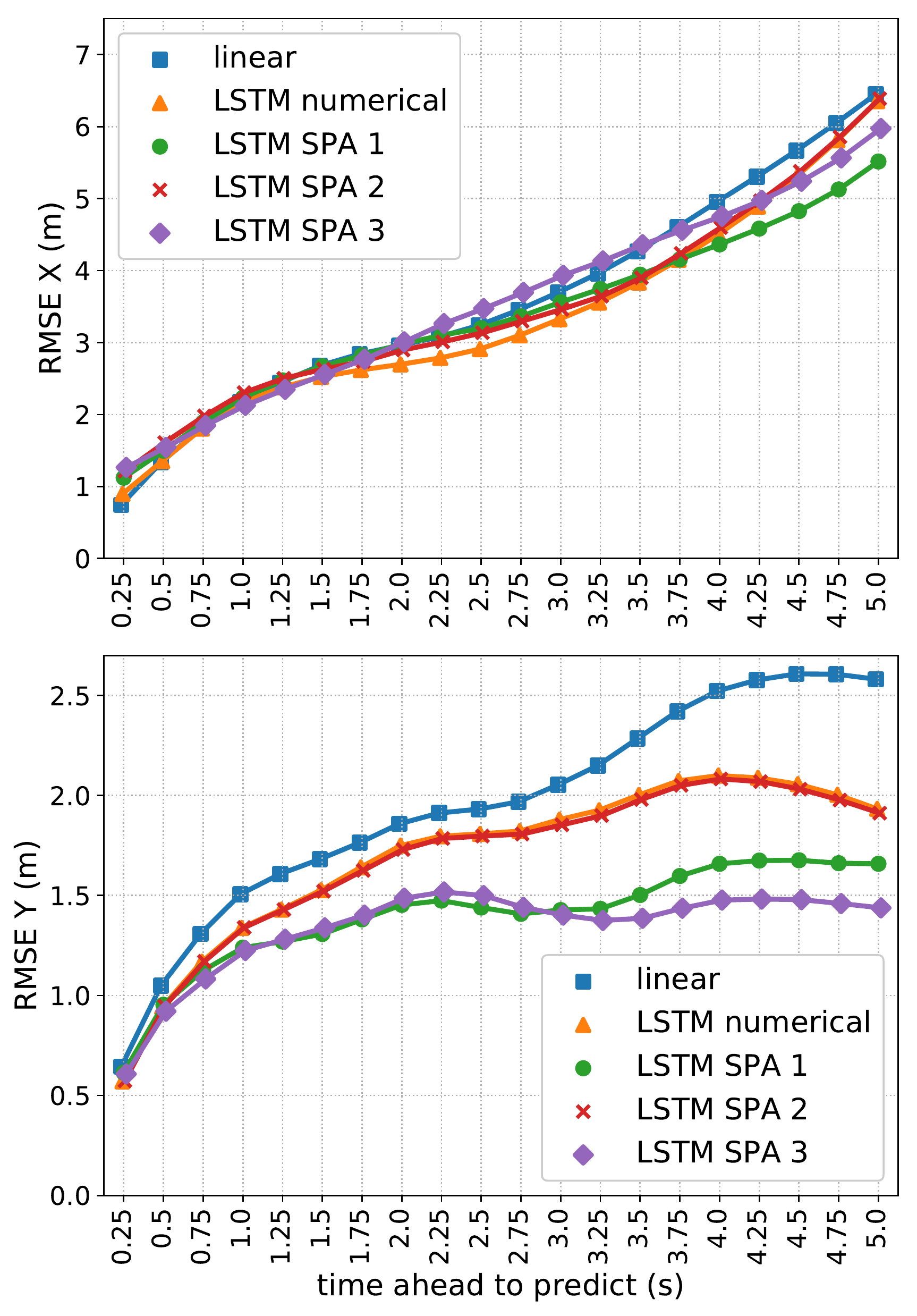}
    }
    \caption{Visualization of the \ac{RMSE} performance of our \ac{LSTM} models for different data setups.
    Figures~\protect\subref{subfig:rmse_large_all},~\protect\subref{subfig:rmse_large_all_lc_only},~\protect\subref{subfig:rmse_large_subset} and~\protect\subref{subfig:rmse_large_subset_lc_only} show the \ac{RMSE} for models trained on the complete data set, while
Fig.~\protect\subref{subfig:rmse_large_all_trained_on_lc},~\protect\subref{subfig:rmse_large_all_lc_only_trained_on_lc},~\protect\subref{subfig:rmse_large_subset_trained_on_lc},~\protect\subref{subfig:rmse_large_subset_lc_only_trained_on_lc} show the same models evaluated on the same samples but trained only on the samples including a target vehicle lane change.
Table~\ref{tab:eval_setups} gives a more detailed overview of the setup for each individual sub-figure.
}
    \label{fig:rmse_on_board_training_data_variation}
\end{figure*}

In this work, we analyze the \acf{LSTM} models used in \cite{Mirus2019b} for the prediction of vehicle positions regarding variations in the data used for training and evaluation.
All networks consist of one \ac{LSTM} encoder and decoder cell for sequence to sequence prediction, which means that the input and the final result of our model is sequential data.
The encoder \ac{LSTM} takes positional data for $20$ past, equidistant time frames as input.
That is, the input data is a sequence of \num{20} items of either positions of the target vehicle or a sequence of high-dimensional vectors encoding this positional data (see Sec.~\ref{subsec:vector_representation} and Table~\ref{tab:evaluated_models} for further details).
Thus, the resulting embedding vector encodes the history of the input data over those \num{20} time frames.
Finally, the embedding vector is used as input for the decoder \ac{LSTM} to predict future vehicle positions.
The output of each model is a sequence of \num{20} positions of the target vehicle predicted over a certain temporal horizon into the future.
The only difference is the dimensionality and encoding of the input data.
Similar to \cite{Mirus2019b}, we use a linear regression model based on a constant velocity assumption as a simple baseline for reference.
Table~\ref{tab:evaluated_models} gives an overview of the models used in this paper.
Further details on implementation and training of the individual models can be found in \cite{Mirus2019b}.

\begin{table}[t]
\begin{center}
\begin{scriptsize}
    \begin{tabular}{|c|c|c|c|c|}
        \hline
        \thead{Short \\ name} & \thead{Input} & \thead{Position \\ encoding} & \thead{Prediction \\ model} & \thead{Data  \\set}\\ \hline
        linear & \makecell{current position \\ and velocity} & - & \makecell{Linear \\ regression} & both \\ \hline
        \makecell{LSTM \\ numerical} & \makecell{sequence \\ of positions} & - & \makecell{\acs{LSTM}} & both \\ \hline
        \makecell{LSTM \\ \acs{SPA} \num{1}} & \makecell{semantic vector \\ sequence} & \makecell{convolutive \\ power} & \makecell{\acs{LSTM}} &both \\ \hline
        \makecell{LSTM \\ \acs{SPA} \num{2}} & \makecell{semantic vector \\ sequence} & \makecell{scalar \\ multiplication} & \makecell{\acs{LSTM}}  &both \\ \hline
        \makecell{LSTM \\ \acs{SPA} \num{3}} & \makecell{semantic vector \\ sequence} & \makecell{convolutive power \\ incl.\ ego-vehicle} & \makecell{\acs{LSTM}}  & \makecell{On-\\ board} \\ \hline
    \end{tabular}
    \vspace{0.2cm}
    \caption{Summary of the employed models regarding learning approach, input data and position encoding.}
	\label{tab:evaluated_models}
\end{scriptsize}
\end{center}
\end{table}

\section{Experiments}%
\label{sec:experiments}

\begin{figure*}[t!]
    \centering
    \subfloat[\label{subfig:rmse_large_all_spa_power_trained_on_lc_vs_trained_on_all}All samples]{%
        \includegraphics[width=0.25\linewidth]{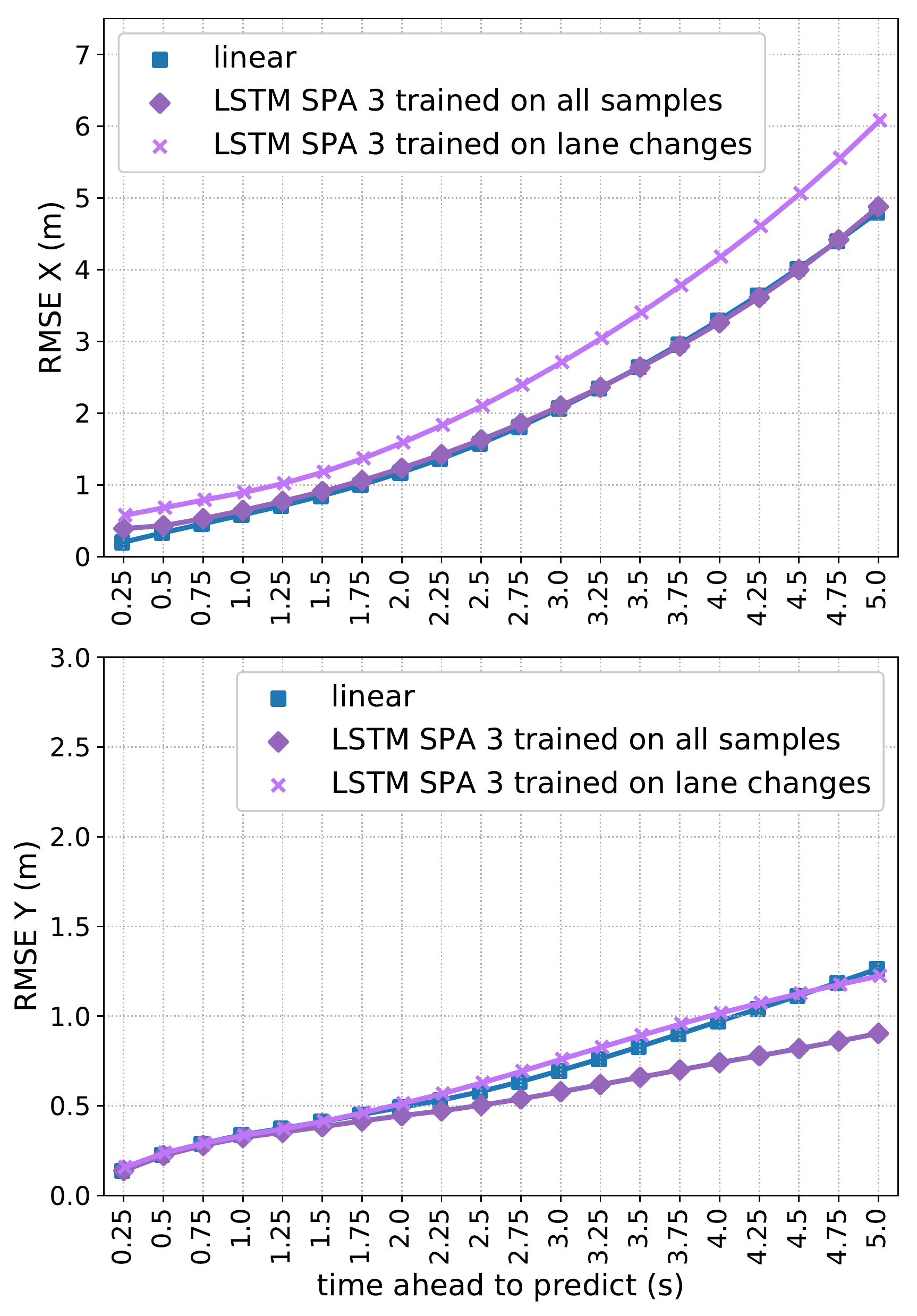}
    }
    \subfloat[\label{subfig:rmse_large_all_lc_only_spa_power_trained_on_lc_vs_trained_on_all}Lane changes]{%
        \includegraphics[width=0.25\linewidth]{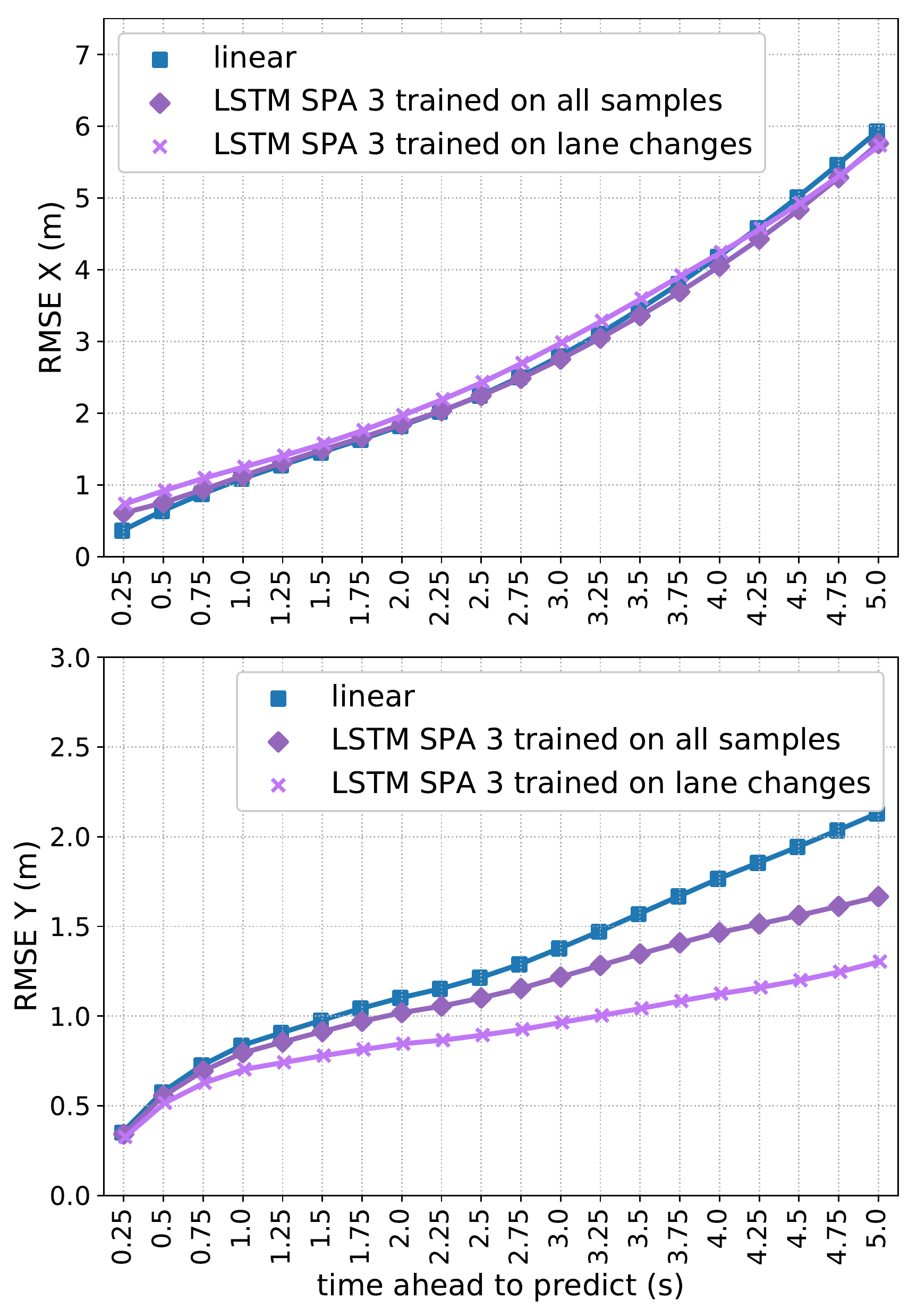}
    }
    \subfloat[\label{subfig:rmse_large_subset_spa_power_trained_on_lc_vs_trained_on_all}Crowded]{%
        \includegraphics[width=0.25\linewidth]{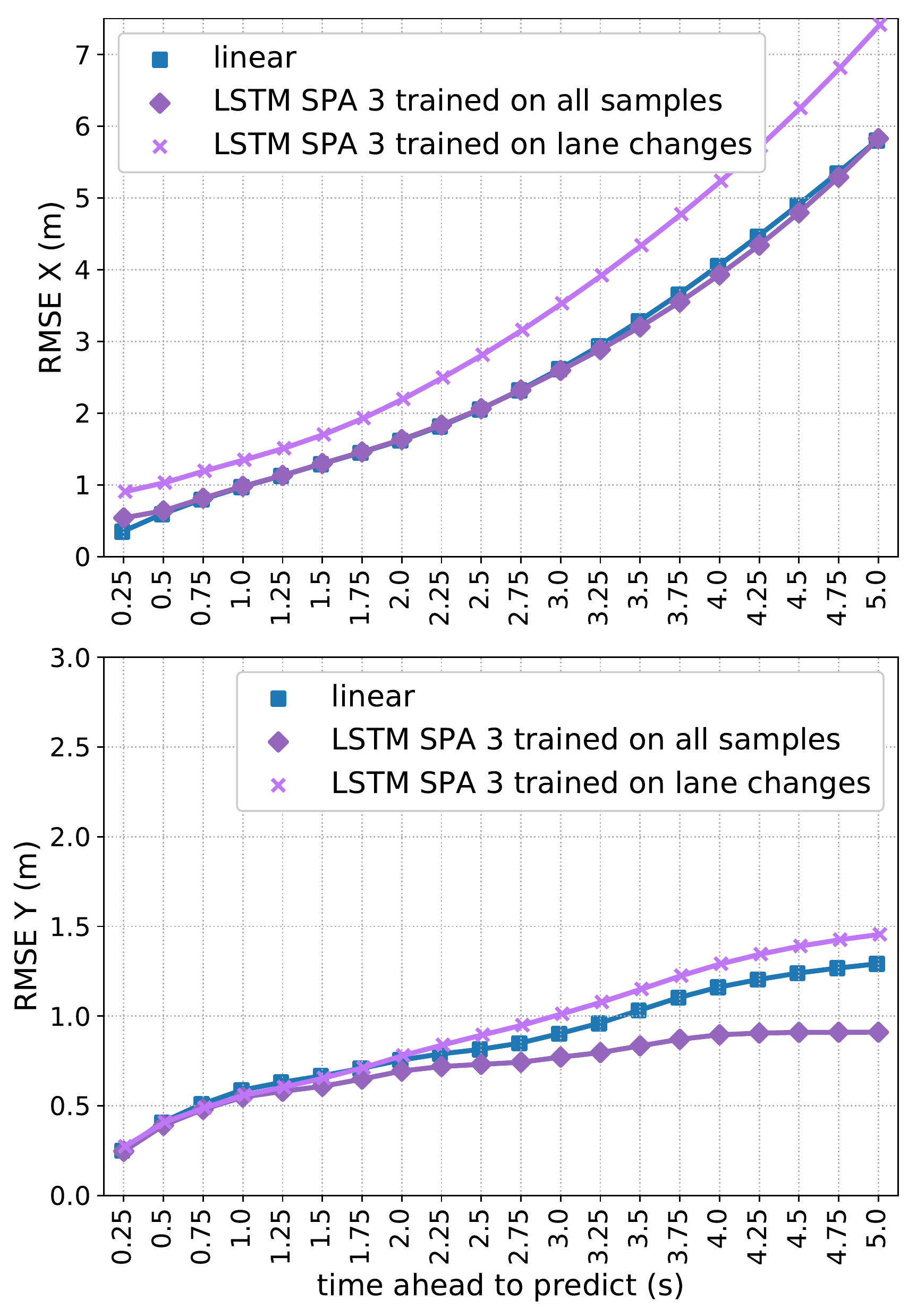}
    }
    \subfloat[\label{subfig:rmse_large_subset_lc_only_spa_power_trained_on_lc_vs_trained_on_all}Crowded lane changes]{%
        \includegraphics[width=0.25\linewidth]{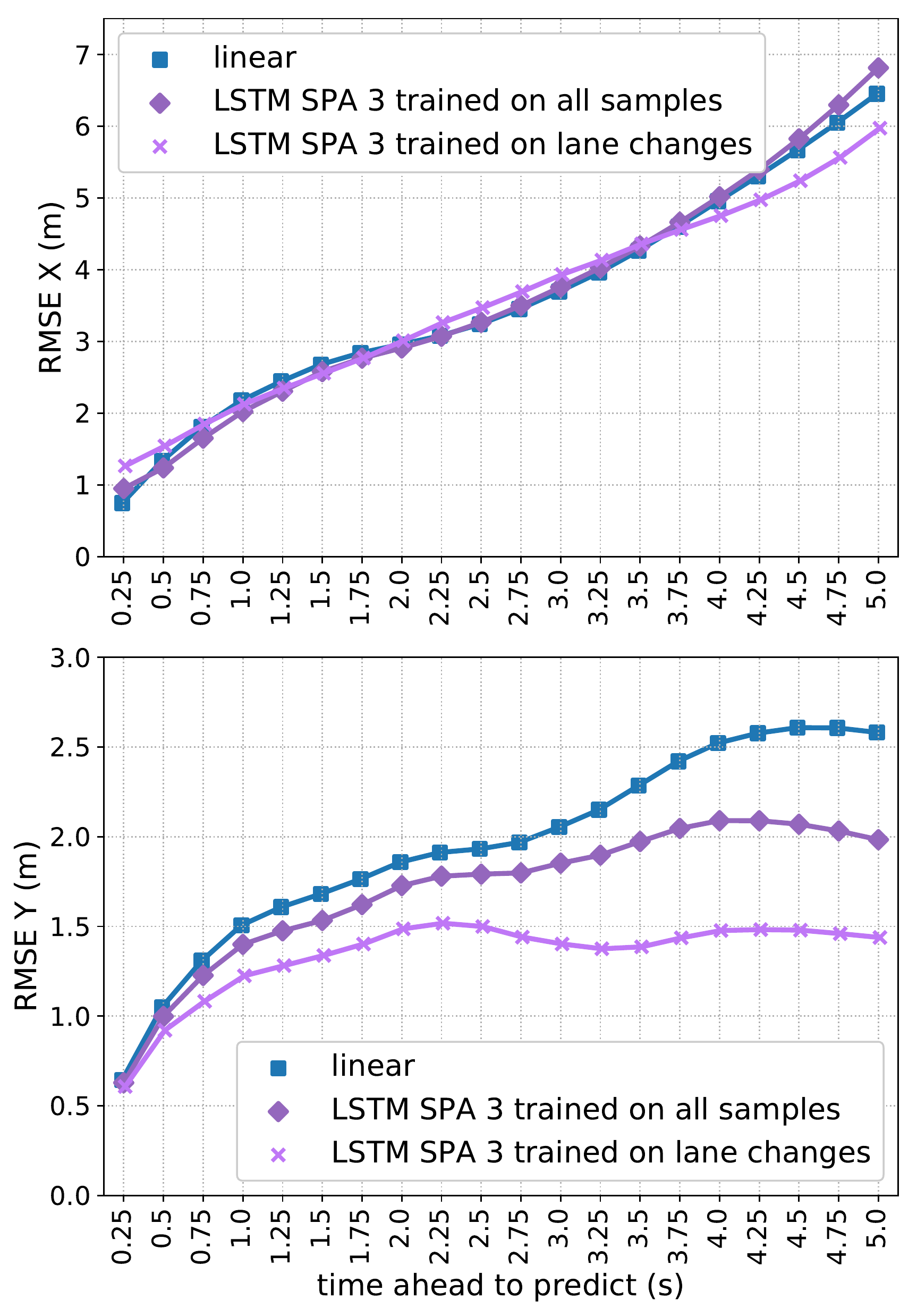}
    }\\
    \subfloat[\label{subfig:rmse_large_all_numerical_trained_on_lc_vs_trained_on_all}All samples]{%
        \includegraphics[width=0.25\linewidth]{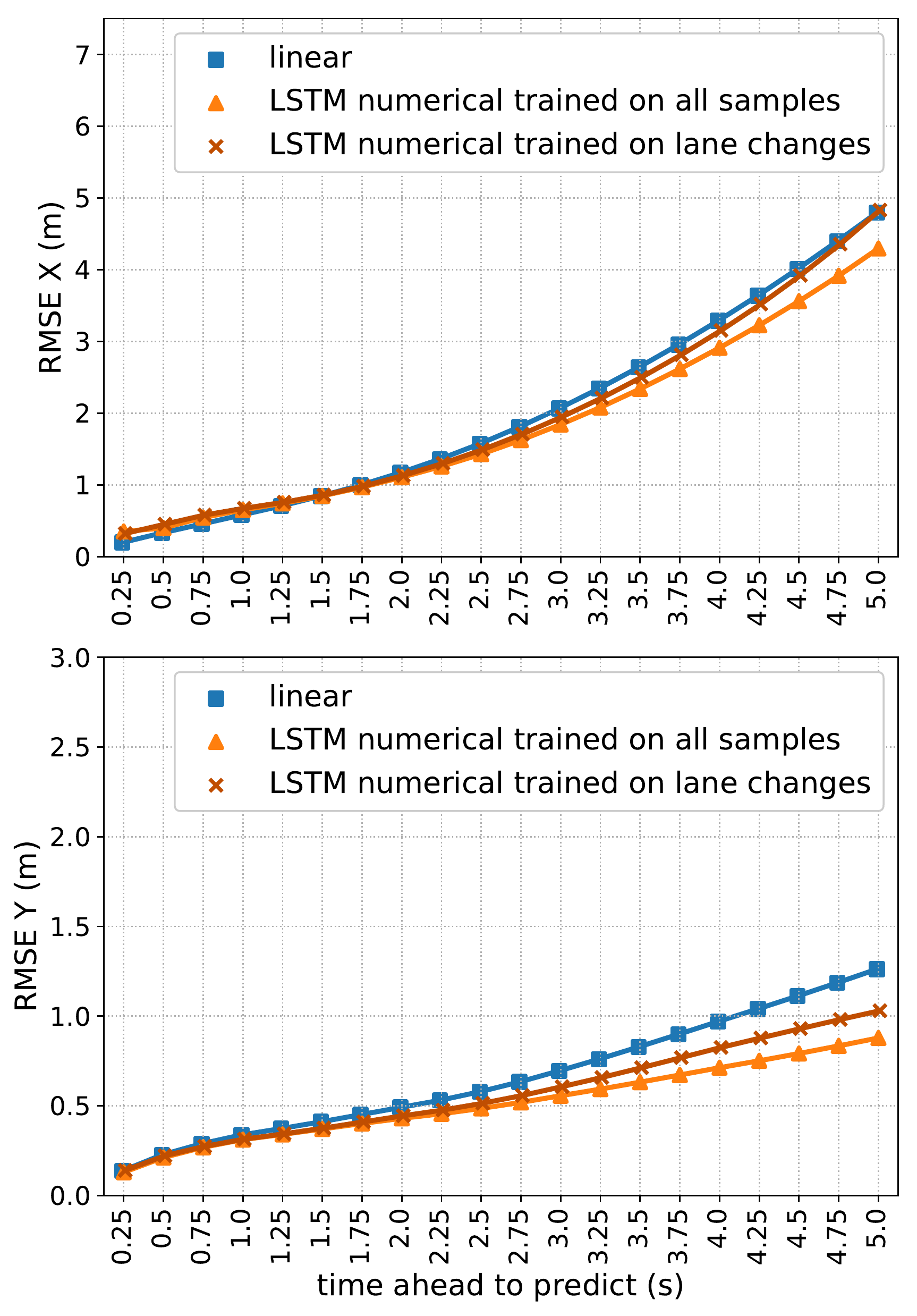}
    }
    \subfloat[\label{subfig:rmse_large_all_lc_only_numerical_trained_on_lc_vs_trained_on_all}Lane changes]{%
        \includegraphics[width=0.25\linewidth]{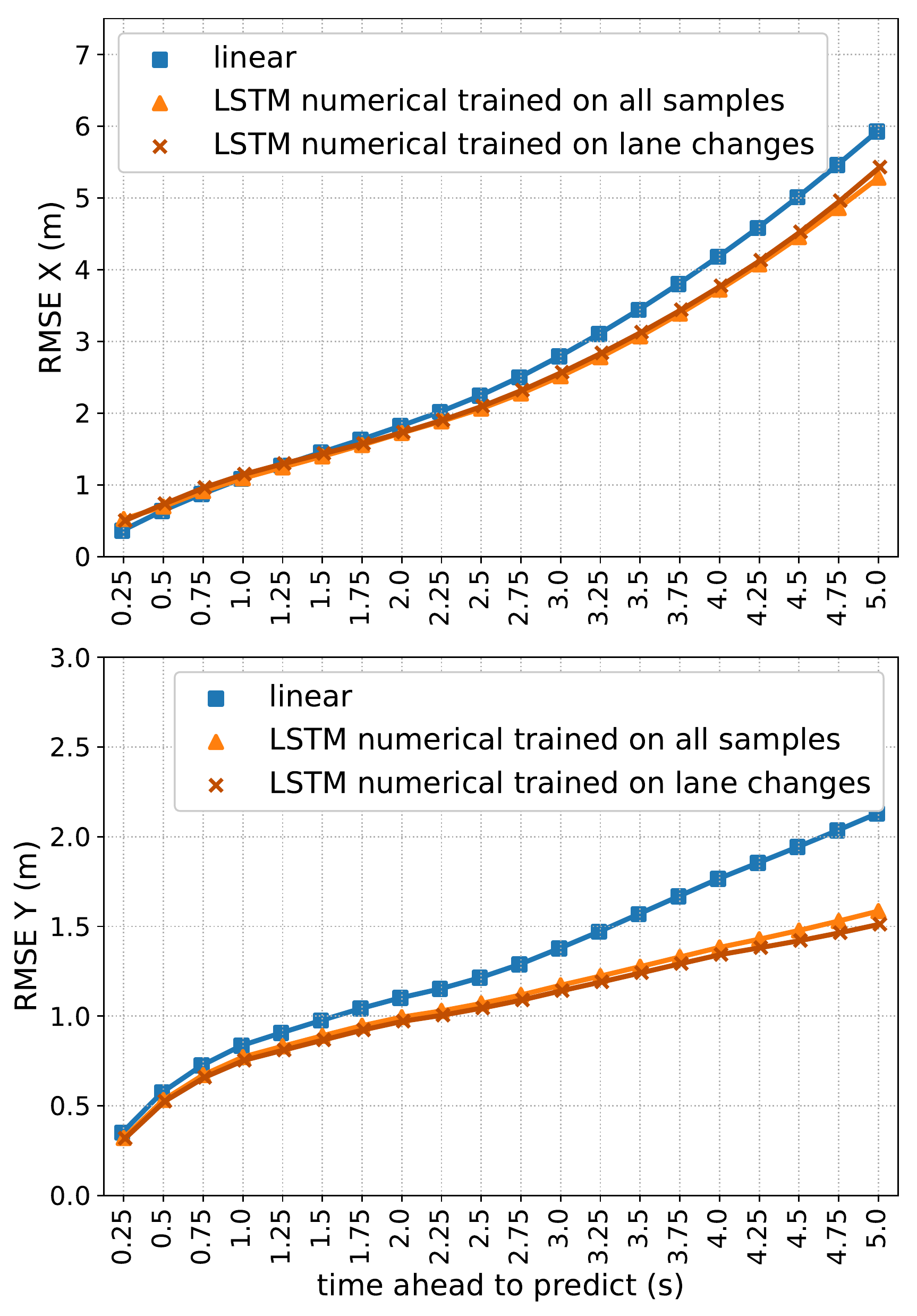}
    }
    \subfloat[\label{subfig:rmse_large_subset_numerical_trained_on_lc_vs_trained_on_all}Crowded]{%
        \includegraphics[width=0.25\linewidth]{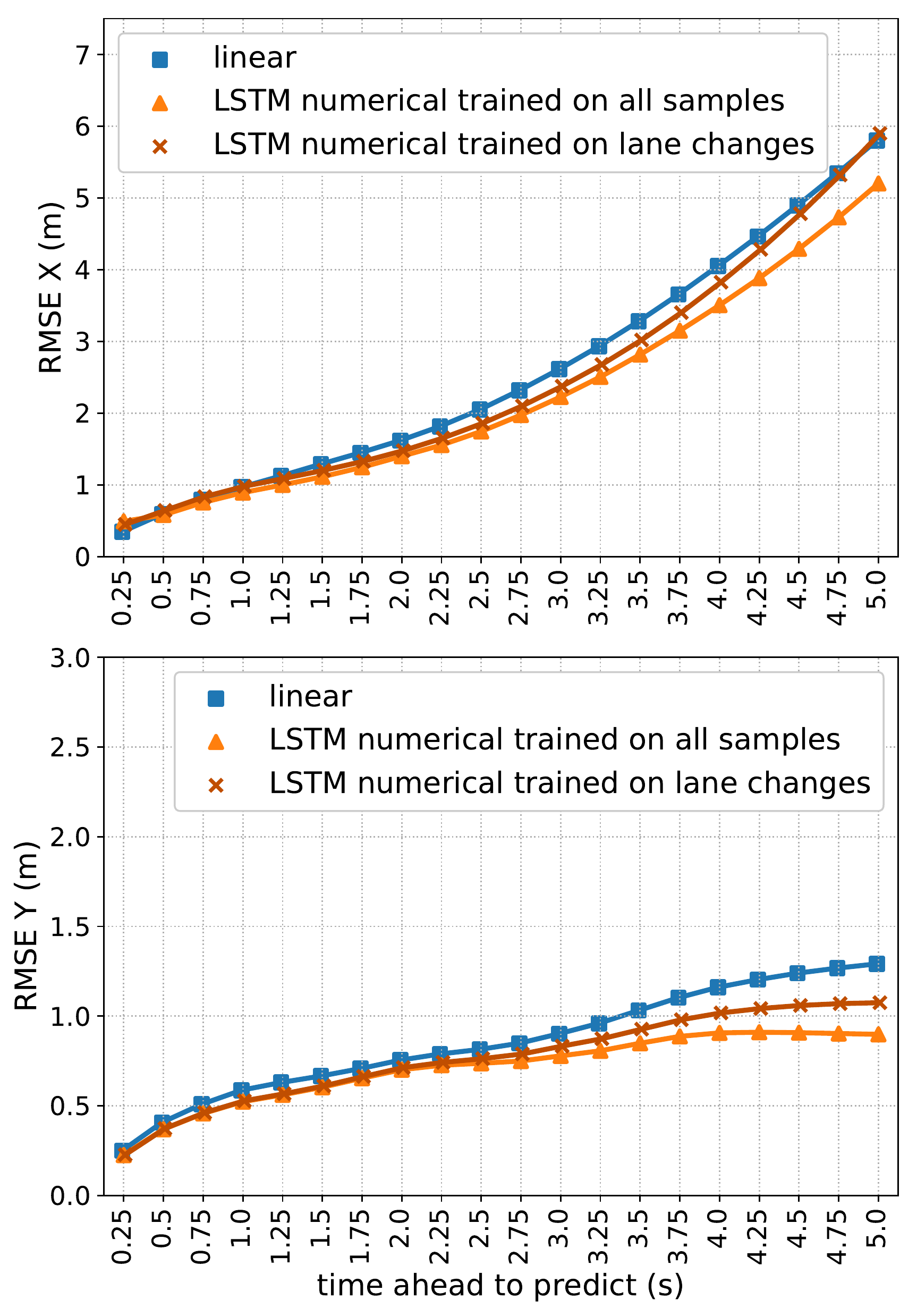}
    }
    \subfloat[\label{subfig:rmse_large_subset_lc_only_numerical_trained_on_lc_vs_trained_on_all}Crowded lane changes]{%
        \includegraphics[width=0.25\linewidth]{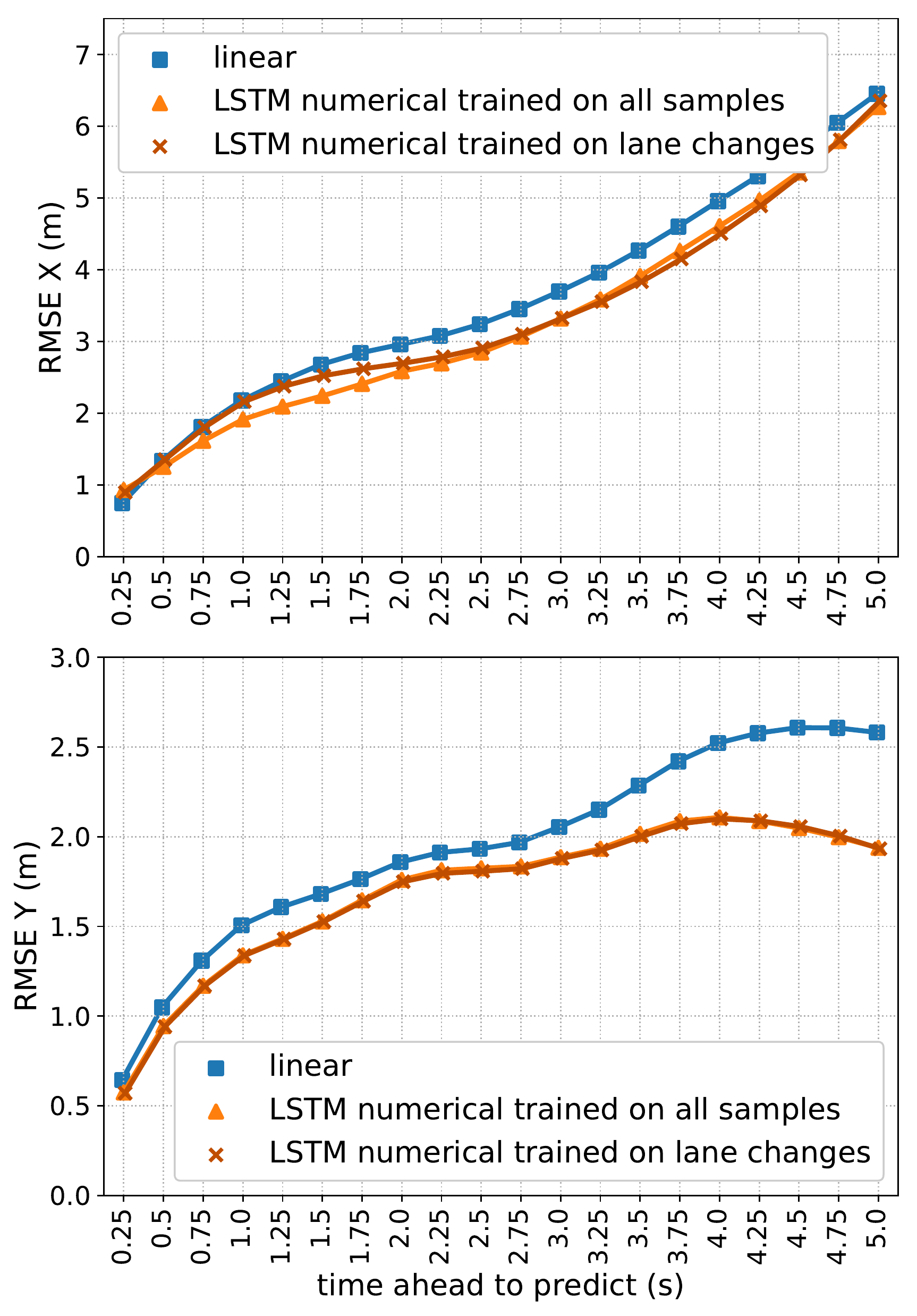}
    }
    \caption{Visualization of the changing \ac{RMSE} performance of particular prediction models depending on the data they have been trained on.
        The first four figures~\protect\subref{subfig:rmse_large_all_spa_power_trained_on_lc_vs_trained_on_all} -~\protect\subref{subfig:rmse_large_subset_lc_only_spa_power_trained_on_lc_vs_trained_on_all} illustrate the difference between the \ac{LSTM} \acs{SPA} \num{3} models when changing their training data, while the last four figures~\protect\subref{subfig:rmse_large_all_numerical_trained_on_lc_vs_trained_on_all} -
        ~\protect\subref{subfig:rmse_large_subset_lc_only_numerical_trained_on_lc_vs_trained_on_all} shows the same comparison for the \ac{LSTM} numerical models.
    }
    \label{fig:rmse_on_board_training_all_vs_training_on_lc_only}
\end{figure*}

In section~\ref{subsec:data_set_composition}, we have already seen that our prediction models using neural networks have to deal with imbalanced data sets containing significantly more straight driving than lane changes performed by the target vehicle.
Hence, training any learning system on all the samples of both our data sets will expose the system to a significantly higher amount of data, in which most likely already a simple linear prediction model performs reasonably well. 
\begin{table}[t]
\begin{center}
\begin{scriptsize}
    \begin{tabular}{|c|c|c|}
        \hline
        \thead{Setup ID} & \thead{Training set} & \thead{Evaluation set} \\ \hline
        (a) & \makecell{all samples} & \makecell{all samples} \\ \hline
        (b) & \makecell{lane change samples} & \makecell{all samples} \\ \hline
        (c) & \makecell{all samples} & \makecell{lane change samples} \\ \hline
        (d) & \makecell{lane change samples} & \makecell{lane change samples} \\ \hline
        (e) & \makecell{all samples} & \makecell{crowded samples} \\ \hline
        (f) & \makecell{lane change samples} & \makecell{crowded samples} \\ \hline
        (g) & \makecell{all samples} & \makecell{crowded lane change samples} \\ \hline
        (h) & \makecell{lane change samples} & \makecell{crowded lane change samples} \\ \hline
    \end{tabular}
    \vspace{0.2cm}
    \caption{Summary of the data samples used for the evaluations shown in individual sub-figures of Fig.~\ref{fig:rmse_on_board_training_data_variation}.}
	\label{tab:eval_setups}
    \vspace{-0.9cm}
\end{scriptsize}
\end{center}
\end{table}
In this section, we therefore analyze the influence of the training data set on our \ac{LSTM} models and if there are significant differences in the performance of models trained on the complete data sets and on subsets consisting only of such samples that contain a lane change of the target vehicle.
We conduct this analysis on the \emph{On-board} data set only.
Figure~\ref{subfig:rmse_large_all} shows the performance of our \ac{LSTM}-based models trained on the complete data set.
Here, we train models with the exact same neural network architectures and encoding schemes of the input just on different data sets, namely the subset of samples containing a lane change performed by the target vehicle.
We consider both types of target vehicle lane changes, namely those performed during the trajectory history and lane changes performed in the future trajectory to be predicted, as input for the models.
Figure~\ref{fig:rmse_on_board_training_data_variation} shows a comparison of different setups regarding training and evaluation data for our \ac{LSTM}-based trajectory prediction models.
Table~\ref{tab:eval_setups} gives the details of the training and evaluation data setups visualized in Fig.~\ref{fig:rmse_on_board_training_data_variation}.
For the term \enquote{crowded samples} as mentioned in Table~\ref{tab:eval_setups}, we refer to the metrics established in \cite{Mirus2019b}: we consider samples with a distance less than \SI{20}{\meter} between the target- and ego-vehicle, with distances less than \SI{20}{\meter} between the target vehicle and the closest other vehicle, and with at least \num{3} other vehicles present in the scene as crowded situations.
Furthermore, Table~\ref{tab:evaluated_models} gives an overview of the different prediction models and their short name used in the legends of Fig.~\ref{fig:rmse_on_board_training_data_variation} and~\ref{fig:rmse_on_board_training_all_vs_training_on_lc_only}.
In both, Fig.~\ref{fig:rmse_on_board_training_data_variation} and~\ref{fig:rmse_on_board_training_all_vs_training_on_lc_only}, we also show a simple linear prediction model based on a constant velocity assumption for reference, since it is not a data-driven learning model and is therefore invariant under the changes of the training data.
Figures~\ref{subfig:rmse_large_all},~\ref{subfig:rmse_large_all_lc_only},~\ref{subfig:rmse_large_subset} and~\ref{subfig:rmse_large_subset_lc_only} show the \ac{RMSE} for models trained on the complete data set while Fig.~\ref{subfig:rmse_large_all_trained_on_lc},~\ref{subfig:rmse_large_all_lc_only_trained_on_lc},~\ref{subfig:rmse_large_subset_trained_on_lc} and~\ref{subfig:rmse_large_subset_lc_only_trained_on_lc} show the same models evaluated on the same samples but trained only on data samples including a target vehicle lane change.
On the other hand, the upper row of Fig.~\ref{fig:rmse_on_board_training_data_variation}, i.e., Fig.~\ref{subfig:rmse_large_all} -~\ref{subfig:rmse_large_all_lc_only_trained_on_lc} illustrates the performance of the models trained on either the complete data set or on only the lane changes evaluated on all data samples, while the lower row, i.e., Fig.~\ref{subfig:rmse_large_subset} -~\ref{subfig:rmse_large_subset_lc_only_trained_on_lc} shows
a similar evaluation for crowded driving situations.

We observe that the models that have been trained only on samples containing a lane change performed by the target vehicle tend to achieve worse results than the models trained on the complete data set, when being evaluated on the entirety of all data samples (Fig.~\ref{subfig:rmse_large_all},~\ref{subfig:rmse_large_subset},~\ref{subfig:rmse_large_all_lc_only} and~\ref{subfig:rmse_large_subset_lc_only}).
Interestingly, the performance of the models based on the convolutive power encoding scheme (\acs{LSTM} \acs{SPA} \num{1} and \num{3}) deteriorates more significantly compared to the other data-driven models, especially in lateral ($y$) direction.
However, if we evaluate the same models only on the samples containing a target vehicle lane change, their performance changes significantly (Fig.~\ref{subfig:rmse_large_all_trained_on_lc},~\ref{subfig:rmse_large_all_lc_only_trained_on_lc},~\ref{subfig:rmse_large_subset_trained_on_lc} and~\ref{subfig:rmse_large_subset_lc_only_trained_on_lc}).
We recapitulate the findings of section~\ref{subsec:data_set_composition} that lane changes influence the lateral ($y$) position values more severely than the longitudinal ($x$) position compared to straight driving samples.
Therefore, the performance difference between the models trained on lane changes only and models trained on the complete data set, as we would expect, is also not that significant in longitudinal direction.
Considering the lateral ($y$) direction however, we observe a significant change between both model and evaluation variants.
If the models are trained only on lane change samples, the \ac{LSTM} \acs{SPA} \num{1} and \num{3} models outperform all other models in lateral direction when evaluated only on the data samples containing a lane change while their performance in longitudinal direction does not change significantly compared to the models trained on the complete data set.

So far, we have only compared either all models trained on the complete data set or all models trained only on the lane change samples.
However, we are also interested in how the performance of particular models changes when modifying the underlying training data set.
Figure~\ref{fig:rmse_on_board_training_all_vs_training_on_lc_only} shows a comparison of the \ac{LSTM}-based models when modifying the training data for the \ac{LSTM} \acs{SPA} \num{3} model (Fig.~\ref{subfig:rmse_large_all_spa_power_trained_on_lc_vs_trained_on_all} -~\ref{subfig:rmse_large_subset_lc_only_spa_power_trained_on_lc_vs_trained_on_all}) as well as the \acs{LSTM} numerical model (Fig.
~\ref{subfig:rmse_large_all_numerical_trained_on_lc_vs_trained_on_all} -~\ref{subfig:rmse_large_subset_lc_only_numerical_trained_on_lc_vs_trained_on_all}).
In this direct comparison, we observe that there is no significant difference in the performance of the numerical \ac{LSTM} models trained on different samples for both evaluation sets containing either the complete data set or only the target vehicle lane changes.
For the \ac{LSTM} \acs{SPA} \num{3} model however, we observe significant improvements for the model trained on the lane change samples when evaluated on the lane changes performed by the target vehicle compared to the model trained on all data samples.
This result could indicate that the numerical model trained on all samples generalizes sufficiently well to all possible situations compared to the convolutive power based model.
However, we have already seen, that both trajectory data sets show a significant imbalance towards straight driving compared to lane change maneuvers (cf.\ section~\ref{subsec:data_set_composition}), which is the same for all models.
Therefore, we believe that the results shown here rather suggest that the learning models employing the convolutive vector-power and thereby encapsulating the prior motion not only of the target vehicle but also other vehicles in its surroundings are better suited to predict lane change maneuvers given a more balanced data set.

\section{Discussion}%
\label{sec:discussion}

\subsection{Conclusion}%
\label{subsec:conclusion}

In this paper, we have analyzed the influence of the composition of training data sets on neural network based models for vehicle trajectory prediction.
In particular, we focused on models employing \acp{LSTM} in combination with a semantic vector representation of automotive scenes and compared them with similar networks employing simple numerical representations of the input data.
We found that when training the \ac{LSTM} models only on data samples containing a lane change performed by the target vehicle, that the model employing the convolutive vector-power representation outperforms all other approaches in $y$-direction when evaluated on the samples containing a lane change, especially in crowded and potentially dangerous situations.
These results suggest that training the whole system on a more balanced data set containing equally many lane change and straight driving samples could improve overall model performance.
Furthermore, the results suggest that the networks using semantic vectors benefit from a more focused and specific training data set, since the models trained particularly on lane changes improved especially in lane change situations compared to network variants trained on all samples.
We expect that similar observations regarding data set imbalance as described in this paper also occur when replacing \acp{LSTM} with  other neural network architectures for sequence to sequence prediction such as \acp{TCN} \cite{Bai2018}, but we leave a more detailed investigation of other network architectures for future work.
Finally, there is not only room for improvement for the models investigated here, but also other data-driven models taking interactions between several vehicles into account used for trajectory prediction in general, by researching and evaluating the influence of the distribution of driving situations within the training data.

\subsection{Future work}%
\label{subsec:future_work}

The results presented in this paper indicate that distributed representations are an interesting option to encode automotive scenes, especially when modeling mutual interactions between traffic participants and using the representational substrate as input to neural networks.
However, one key strength of such representations and therefore essential for them to unfold their full potential, is the possibility of implementation in \acp{SNN} \cite{Eliasmith2013} and therefore deployment on dedicated neuromorphic hardware.
This deployment can only be achieved in combination with prediction models employing a spiking neuron substrate as well, which could be an interesting, energy-efficient option in future automated vehicles with tight restrictions regarding on-board power consumption.
Unfortunately, \ac{LSTM}-based neural networks neither allow implementation in spiking neurons nor deployment on neuromorphic hardware.
Therefore, one direction for future research could be to replace the \acp{LSTM} in our current approach with the recently proposed \acp{LMU} \cite{Voelker2019}, which offer prediction capabilities at least similar to or even better than \acp{LSTM}, but also allow implementation in \acp{SNN} and therefore deployment on neuromorphic hardware.

Another interesting option for future research is to investigate the performance of our prediction models on other publicly available data sets with particular focus on the composition of the data set.
Apart from the imbalanced \emph{\ac{NGSIM}} data set, possible candidates for this investigation could be the recently proposed INTERACTION data set \cite{Zhan2019} or the trajectory section of Baidu's ApolloScapes data set \cite{Huang2018}.
On the other hand, one could investigate situations other than lane changes such as sudden acceleration or braking, which could also be interesting for learning models to predict, or to further balance the training data of such models.
Finally, both data sets used in this work contain mainly highway driving samples.
When moving to more advanced data sets containing urban and interurban driving with a far larger variety of possible scenarios and situations, the question of how to properly balance a trajectory prediction data set for automated driving becomes even more interesting but also more challenging.

\IEEEtriggeratref{13}

\bibliographystyle{IEEEtran}
\bibliography{literature}

\begin{thebibliography}{10}
\providecommand{\url}[1]{#1}
\csname url@samestyle\endcsname
\providecommand{\newblock}{\relax}
\providecommand{\bibinfo}[2]{#2}
\providecommand{\BIBentrySTDinterwordspacing}{\spaceskip=0pt\relax}
\providecommand{\BIBentryALTinterwordstretchfactor}{4}
\providecommand{\BIBentryALTinterwordspacing}{\spaceskip=\fontdimen2\font plus
\BIBentryALTinterwordstretchfactor\fontdimen3\font minus
  \fontdimen4\font\relax}
\providecommand{\BIBforeignlanguage}[2]{{%
\expandafter\ifx\csname l@#1\endcsname\relax
\typeout{** WARNING: IEEEtran.bst: No hyphenation pattern has been}%
\typeout{** loaded for the language `#1'. Using the pattern for}%
\typeout{** the default language instead.}%
\else
\language=\csname l@#1\endcsname
\fi
#2}}
\providecommand{\BIBdecl}{\relax}
\BIBdecl

\bibitem{Ciresan2012}
D.~C. Ciresan, U.~Meier, J.~Masci, and J.~Schmidhuber, ``Multi-column deep
  neural network for traffic sign classification,'' \emph{Neural Networks},
  vol.~32, pp. 333--338, 2012.

\bibitem{Bojarski2016}
\BIBentryALTinterwordspacing
M.~Bojarski, D.~D. Testa, D.~Dworakowski, B.~Firner, B.~Flepp, P.~Goyal, L.~D.
  Jackel, M.~Monfort, U.~Muller, J.~Zhang, X.~Zhang, J.~Zhao, and K.~Zieba,
  ``End to end learning for self-driving cars,'' \emph{arXiv, Computing
  Research Repository (CoRR)}, vol. abs/1604.07316, 2016. [Online]. Available:
  \url{http://arxiv.org/abs/1604.07316}
\BIBentrySTDinterwordspacing

\bibitem{Lawitzky2013}
A.~Lawitzky, D.~Althoff, C.~F. Passenberg, G.~Tanzmeister, D.~Wollherr, and
  M.~Buss, ``Interactive scene prediction for automotive applications,'' in
  \emph{Intelligent Vehicles Symposium (IV), 2013 IEEE}, 2013, pp. 1028--1033.

\bibitem{Lefevre2014}
S.~Lef{\`e}vre, D.~Vasquez, and C.~Laugier, ``A survey on motion prediction and
  risk assessment for intelligent vehicles,'' \emph{ROBOMECH Journal}, vol.~1,
  no.~1, p.~1, 2014.

\bibitem{Polychronopoulos2007}
A.~Polychronopoulos, M.~Tsogas, A.~Amditis, and L.~Andreone, ``Sensor fusion
  for predicting vehicles' path for collision avoidance systems,'' \emph{{IEEE}
  Transactions on Intelligent Transportation Systems}, vol.~8, no.~3, pp.
  549--562, 2007.

\bibitem{Schmuedderich2015}
J.~Schm{\"u}dderich, S.~Rebhan, T.~Weisswange, M.~Kleinehagenbrock, R.~Kastner,
  M.~Nishigaki, S.~Kusuhara, H.~Kamiya, N.~Mori, and S.~Ishida, ``A novel
  approach to driver behavior prediction using scene context and physical
  evidence for intelligent adaptive cruise control (i-acc),'' in \emph{Future
  Active Safety Technology Towards zero traffic accidents (FAST-zero)}.\hskip
  1em plus 0.5em minus 0.4em\relax FISITA, 2015.

\bibitem{Bahram2016}
M.~Bahram, C.~Hubmann, A.~Lawitzky, M.~Aeberhard, and D.~Wollherr, ``A
  {C}ombined {M}odel- and {L}earning-{B}ased {F}ramework for
  {I}nteraction-{A}ware {M}aneuver {P}rediction,'' \emph{IEEE Transactions on
  Intelligent Transportation Systems}, vol.~17, no.~6, pp. 1538--1550, 2016.

\bibitem{Bonnin2012}
S.~Bonnin, F.~Kummert, and J.~Schm{\"u}dderich, ``A generic concept of a system
  for predicting driving behaviors,'' in \emph{2012 15th International IEEE
  Conference on Intelligent Transportation Systems}, 2012, pp. 1803--1808.

\bibitem{Hochreiter1997}
S.~Hochreiter and J.~Schmidhuber, ``Long short-term memory,'' \emph{Neural
  Computation}, vol.~9, no.~8, pp. 1735--1780, nov 1997.

\bibitem{Alahi2016}
A.~Alahi, K.~Goel, V.~Ramanathan, A.~Robicquet, L.~Fei-Fei, and S.~Savarese,
  ``Social {LSTM}: {H}uman {T}rajectory {P}rediction in {C}rowded {S}paces,''
  in \emph{2016 IEEE Conference on Computer Vision and Pattern Recognition
  (CVPR)}, 2016, pp. 961--971.

\bibitem{Deo2018a}
\BIBentryALTinterwordspacing
N.~Deo and M.~M. Trivedi, ``Convolutional social pooling for vehicle trajectory
  prediction,'' \emph{The IEEE Conference on Computer Vision and Pattern
  Recognition (CVPR) Workshops, 2018, pp. 1468-1476}, 2018. [Online].
  Available: \url{http://arxiv.org/abs/1805.06771}
\BIBentrySTDinterwordspacing

\bibitem{Altche2018}
\BIBentryALTinterwordspacing
F.~Altche and A.~de~La~Fortelle, ``An {LSTM} network for highway trajectory
  prediction,'' in \emph{2017 IEEE 20th International Conference on Intelligent
  Transportation Systems (ITSC)}.\hskip 1em plus 0.5em minus 0.4em\relax
  {IEEE}, Oct 2017, pp. 353--359. [Online]. Available:
  \url{http://dx.doi.org/10.1109/itsc.2017.8317913}
\BIBentrySTDinterwordspacing

\bibitem{Deo2018}
N.~Deo and M.~M. Trivedi, ``Multi-modal trajectory prediction of surrounding
  vehicles with maneuver based {LSTMs},'' in \emph{2018 {IEEE} Intelligent
  Vehicles Symposium ({IV})}.\hskip 1em plus 0.5em minus 0.4em\relax {IEEE},
  2018, pp. 1179--1184.

\bibitem{Deng2009}
J.~Deng, W.~Dong, R.~Socher, L.-J. Li, K.~Li, and L.~Fei-Fei, ``{ImageNet: A
  Large-Scale Hierarchical Image Database},'' in \emph{IEEE Computer Vision and
  Pattern Recognition (CVPR)}, 2009, pp. 248--255.

\bibitem{Lin2014}
T.-Y. Lin, M.~Maire, S.~Belongie, L.~Bourdev, R.~Girshick, J.~Hays, P.~Perona,
  D.~Ramanan, C.~L. Zitnick, and P.~Dollár, ``Microsoft {COCO}: Common objects
  in context,'' \emph{arXiv e-prints}, p. arXiv:1405.0312, 2014.

\bibitem{Geiger2013a}
A.~Geiger, P.~Lenz, C.~Stiller, and R.~Urtasun, ``Vision meets robotics: The
  {KITTI} dataset,'' \emph{The International Journal of Robotics Research},
  vol.~32, no.~11, pp. 1231--1237, 2013.

\bibitem{Cordts2016}
\BIBentryALTinterwordspacing
M.~Cordts, M.~Omran, S.~Ramos, T.~Rehfeld, M.~Enzweiler, R.~Benenson,
  U.~Franke, S.~Roth, and B.~Schiele, ``The {C}ityscapes {D}ataset for
  {S}emantic {U}rban {S}cene {U}nderstanding,'' \emph{arXiv, Computing Research
  Repository (CoRR)}, vol. abs/1604.01685, 2016. [Online]. Available:
  \url{http://arxiv.org/abs/1604.01685}
\BIBentrySTDinterwordspacing

\bibitem{Huang2018}
X.~Huang, X.~Cheng, Q.~Geng, B.~Cao, D.~Zhou, P.~Wang, Y.~Lin, and R.~Yang,
  ``The {ApolloScape} dataset for autonomous driving,'' in \emph{The IEEE
  Conference on Computer Vision and Pattern Recognition (CVPR) Workshops},
  2018, pp. 954--960.

\bibitem{Mirus2019b}
F.~Mirus, P.~Blouw, T.~C. Stewart, and J.~Conradt, ``An investigation of
  vehicle behavior prediction using a vector power representation to encode
  spatial positions of multiple objects and neural networks,'' \emph{Frontiers
  in Neurorobotics}, vol.~13, p.~84, oct 2019.

\bibitem{NGSIM-US101}
\BIBentryALTinterwordspacing
J.~Colyar and J.~Halkias, ``{US Highway 101 Dataset},'' 2017. [Online].
  Available:
  \url{https://www.fhwa.dot.gov/publications/research/operations/07030/index.cfm}
\BIBentrySTDinterwordspacing

\bibitem{Zhan2019}
\BIBentryALTinterwordspacing
W.~{Zhan}, L.~{Sun}, D.~{Wang}, H.~{Shi}, A.~{Clausse}, M.~{Naumann},
  J.~{Kummerle}, H.~{Konigshof}, C.~{Stiller}, A.~{de La Fortelle}, and
  M.~{Tomizuka}, ``{INTERACTION Dataset: An INTERnational, Adversarial and
  Cooperative moTION Dataset in Interactive Driving Scenarios with Semantic
  Maps},'' \emph{arXiv e-prints}, p. arXiv:1910.03088, Sep 2019. [Online].
  Available: \url{http://arxiv.org/abs/1910.03088}
\BIBentrySTDinterwordspacing

\bibitem{Eliasmith2013}
C.~Eliasmith, \emph{{H}ow to build a brain: {A} neural architecture for
  biological cognition}.\hskip 1em plus 0.5em minus 0.4em\relax Oxford
  University Press, 2013.

\bibitem{Gayler2003}
R.~Gayler, ``Vector {S}ymbolic {A}rchitectures answer {J}ackendoff's challenges
  for cognitive neuroscience,'' in \emph{ICCS/ASCS International Conference on
  Cognitive Science}, P.~Slezak, Ed., University of New South Wales.\hskip 1em
  plus 0.5em minus 0.4em\relax CogPrints, 2003, pp. 133--138.

\bibitem{Plate1994}
T.~Plate, ``Distributed {R}epresentations and {N}ested {C}ompositional
  {S}tructure,'' Ph.D. dissertation, University of Toronto, 1994.

\bibitem{Bai2018}
\BIBentryALTinterwordspacing
S.~Bai, J.~Z. Kolter, and V.~Koltun, ``An empirical evaluation of generic
  convolutional and recurrent networks for sequence modeling,'' \emph{arXiv,
  Computing Research Repository (CoRR)}, vol. abs/1803.01271, 2018. [Online].
  Available: \url{http://arxiv.org/abs/1803.01271}
\BIBentrySTDinterwordspacing

\bibitem{Voelker2019}
\BIBentryALTinterwordspacing
A.~Voelker, I.~Kaji\'{c}, and C.~Eliasmith, ``{Legendre Memory Units:
  Continuous-Time Representation in Recurrent Neural Networks},'' in
  \emph{Advances in Neural Information Processing Systems 32}, H.~Wallach,
  H.~Larochelle, A.~Beygelzimer, F.~d\textquotesingle Alch\'{e}-Buc, E.~Fox,
  and R.~Garnett, Eds.\hskip 1em plus 0.5em minus 0.4em\relax Curran
  Associates, Inc., pp. 15\,544--15\,553. [Online]. Available:
  \url{http://papers.nips.cc/paper/9689-legendre-memory-units-continuous-time-representation-in-recurrent-neural-networks.pdf}
\BIBentrySTDinterwordspacing

\end{thebibliography}

\end{document}